%% file: conf.tex
\documentclass[letterpaper, 10 pt, conference]{ieeeconf}
\IEEEoverridecommandlockouts

\usepackage[letterpaper, left=0.60in, right=0.85in, bottom=0.645in, top=0.77in]{geometry}

\usepackage{pgfplots}
\pgfplotsset{compat=newest}
\usepackage{tikz}
\usetikzlibrary{calc}
\usetikzlibrary{shapes.geometric, arrows}
\usetikzlibrary{arrows.meta}
\usetikzlibrary{matrix}

\usepackage{graphicx}
\usepackage{caption}
\usepackage{subcaption}
\usepackage{amsmath}
\usepackage{amssymb}
\usepackage{bm}

\usepackage{soul}

\usepackage[noadjust]{cite}
\usepackage{balance}

\usepackage{renew}
\usepackage{mymath,diffge,gmech,lmech}
\usepackage{snakey_tip}

\title{In-Place Rotation for Enhancing Snake-like Robot Mobility}

\author{Alexander H. Chang$^{1}$ and Patricio A. Vela$^{1}$%
\thanks{$1$: The authors are with the School of Electrical and Computer
Engineering, Georgia Institute of Technology, Atlanta, GA, USA.
Email: alexander.h.chang@gatech.edu, pvela@ece.gatech.edu}%
\thanks{This work was supported by NSF grant \#1562911.}
}

\begin{document}
%

\maketitle

\begin{abstract}
Gaits engineered for snake-like robots to rotate in-place instrumentally
fill a gap in the set of locomotive gaits that have traditionally prioritized
translation. This paper designs a Turn-in-Place gait and demonstrates
the ability of a shape-centric modeling framework to capture the gait's
locomotive properties.  Shape modeling for turning involves a time-varying
continuous body curve described by a standing wave. 
Presumed viscous robot-ground frictional interactions lead to body
dynamics conditioned on the time-varying shape model. 
The dynamic equations describing the Turn-in-Place gait are validated
by an articulated snake-like robot using a physics-based simulator and a
physical robot. 
The results affirm the shape-centric modeling framework's capacity 
to model a variety of snake-like robot gaits with fundamentally different
body-ground contact patterns. 
As an applied demonstration, example locomotion scenarios partner the
shape-centric Turn-in-Place gait with a Rectilinear gait for maneuvering
through constrained environments based on a multi-modal locomotive
planning strategy.  Unified shape-centric modeling facilitates
trajectory planning and tracking for a snake-like robot to successfully
negotiate non-trivial obstacle configurations.
\end{abstract}



\section{Introduction}
\seclabel{intro}
\input{intro.tex}

\section{Turn-in-Place Gait Kinematics}
\seclabel{tip_gait_shape}
\input{tip_gait.tex}

\section{Locomotive Dynamics}
\seclabel{dynamics}
\input{dynamics.tex}


\section{Multi-Gait Trajectory Planning and Tracking}
\seclabel{plan_track}
\input{plan_track.tex}

\section{Locomotion Scenarios}
\seclabel{results}
\input{results.tex}

\section{Conclusion}
\seclabel{conclusion}
\input{conclusion.tex}

\balance 
\bibliographystyle{IEEEtran}
\bibliography{IEEEabrv,shorten,robot,robosnake,biomimetic,ivalab}

\end{document}

%% file: intro.tex
Gait-based snake-like robot locomotion over flat ground has primarily
focused on a handful of motion primitives: lateral undulation,
sidewinding and rectilinear forms of motion. These gaits exhibit
predominantly-translational movement.
In application, they are used by snake-like robots to achieve large
translational displacements (many body lengths) to reach prescribed goal
positions. Steering during locomotion involves modulating gait parameters
to induce steering effects
\cite{KoEtAl_BnB[2016], YaWaSh_AIM[2019], AsEtAl_PNAS[2015],
XiEtAl_ICRA[2015], ChVe_RAS[2020]}. For lateral undulation and rectilinear
gaits, steering relies on joint offsets. For conical sidewinding, the  
cone half-angle mediates the turning rate.
The steering parameters govern the robot's body shape during locomotion
and are subject to saturation limits. Beyond the limits, the prescribed
body shape is not realizable, either due to actuator constraints or
due to self-collision.  Consequently, the curved trajectories resulting
from these gaits are similarly subject to a saturating turning rate.
There is a limit to how tightly the robot may turn under
translation-dominant gaits. 

Locomotive models for translational gaits have been derived as either
dynamic or kinematic motion models. Derivations of lateral undulation
and rectilinear models are specialized to the particular gait
of interest and to the underlying snake-like robot design 
\cite{LiEtAl_Springer[2013],TaReMa_IROS[2015],TaMa_ICRA[2015]}. 
Follow-on trajectory control strategies are tied to the particular
gait's motion model, possibly with provable tracking guarantees
\cite{LiEtAl_Springer[2013]}. Conical forms of sidewinding have been
studied under the presumption of ideally-static rolling body-ground
contact. A kinematic analysis results in a closed-form motion model,
parametrized with respect to the gait's input shape parameters
\cite{GoEtAl_ICRA[2012]}. A re-parametrization allows this gait to be
re-framed as a simple differential-drive vehicle, for which traditional
path tracking approaches may then be applied \cite{XiEtAl_ICRA[2015]}.
Alternatively, a shape-centric modeling framework has been presented,
whereby shape kinematics and locomotive dynamics of various bio-inspired
gaits for snake-like robots are described through a unified modeling
approach \cite{ChVe_RAS[2020], ChSeVe_CDC[2016b]}.
Reduced-order motion models, that capture the time-averaged kinematic
behavior of each gait, can be empirically-derived from the full system
dynamics. Reduction of complex dynamics associated with snake-like
robotic gaits, to simplified motion models, facilitates locomotive
planning and control \cite{ChVe_RAS[2020], ChEtAl_ACC[2018]}. 
This model reduction approach realizes the framework proposed in
\cite{McOs_IJRR[2003]}. 
An attractive property of the shape-centric modeling framework is its
ability describe a variety of gaits and to produce uniformly structured
motion models for each. Planning and control strategies 
viable for one gait apply to other gaits modeled using this approach.
This generalization is meaningful for snake-like robots, and potentially
other bio-inspired robots; organisms that serve as bio-inspirational
focus frequently exhibit several modes of locomotion to accommodate
varying environmental situations \cite{LoEtAl_BnB[2010],LoEtAl_BnB[2015]}. 

More recently, gaits for elongated-body, limbless robots have been
designed that `rotate-in-place'. While no known snake species exhibit
gaits that accomplish this motion profile, biological inspiration has
been found in nematode worms, a different elongated-body and limbless
organism \cite{WaEtAl_IROS[2020]}. A turn-in-place gait variant,
designated `frequency turning', has been discovered through an empirical
search through the robot shape space \cite{GoEtAl_ICRA[2015], AsEtAl_PNAS[2015]}. 
The net change in orientation achieved over a single cycle of these gaits 
is large, while net translation remains relatively small. 
Interestingly, the nematode worm-inspired Omega Turn relies solely on
planar shape changes, with no vertical lift, to accomplish rotation;
the latter frequency turning gait instead accomplishes this through
strategic body-ground contact planning. Motion models for these gaits
are grounded in the articulated platforms on which they are applied and
involve elements of empirical sampling for different locomotion
environments.

The unique motion profile characterizing turn-in-place gaits fills an important 
gap in the space of locomotive behaviors available to snake-like robots. 
Additing a turn-in-place gait to their toolbox realizes the ability to
mimic differential-drive wheeled robots. 
As a consequence, planning and tracking control approaches frequently
used on wheeled robots become more comprehensively applicable to
snake-like robots.  These wheeled platforms can steer while commanding
non-zero linear velocities, and can arbitrarily re-orient in-place at
(near) zero linear velocity.  In a similar manner, multi-modal switching
between turn-in-place and translationally-focused gaits permit
snake-like robots to traverse paths that neither gait alone may achieve.
This greatly broadens set of navegable locomotion scenarios for
snake-like robots, as candidate trajectory solutions are no longer
constrained by turning rate limits.


\noindent {\bf Contribution.}
We design a Turn-in-Place gait for snake-like robots with standing wave
shape kinematics. A shape-centric modeling framework
\cite{ChVe_RAS[2020]} conditioned on the gait describes the robot's locomotive
dynamics. 
The gait is represented by a time-varying continuous body curve defined
with respect to a static average body reference curve and rigid body frame. 
Planar group dynamics are derived for this continuous-body gait after
introducing a strategically-engineered body-ground contact pattern and
using a viscous friction model. 
The equations of motion are validated in a physics-based simulator
(Gazebo) and on a physical snake-like robot.
The validation highlights the shape-centric modeling framework's
capacity to describe a variety of snake-like robot gaits in a unified
manner. The uniformly-structured gait models that result are useful in
locomotion planning and control.  
Integrated use of the Turn-in-Place gait with a pre-existing Rectilinear gait 
leads to a multi-gait trajectory synthesis and tracking strategy.
A snake-like robot can then traverse obstacle-strewn scenarios requiring
tight turns by exploiting each gait's affordances.

%% file: tip_gait.tex
\begin{figure*} [!t]
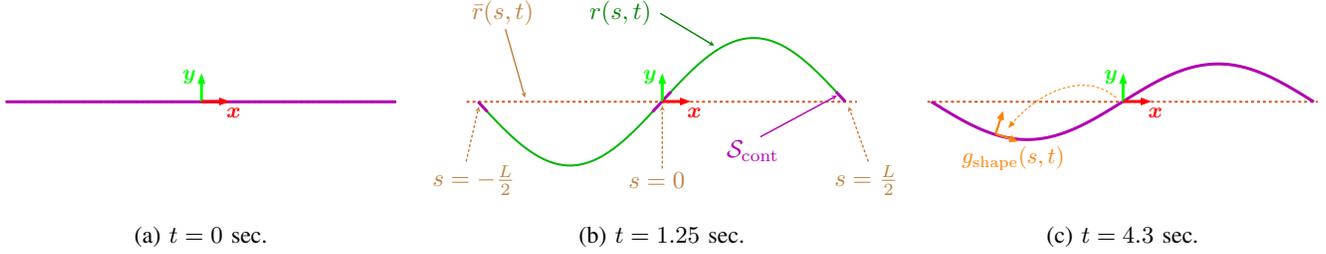

\vspace{1.0mm}
\centering
  \begin{subfigure}{0.65\columnwidth}
  \centering
  \resizebox{1.0\columnwidth}{!}{\input{tip_gait_shape_1.tex}} 
	\caption{$t = 0$ sec.}
	\label{tip_gait_shape01}
	\end{subfigure}
	\hfill
  \begin{subfigure}{0.65\columnwidth}
  \centering
  \resizebox{1.0\columnwidth}{!}{\input{tip_gait_shape_2.tex}} 
	\caption{$t = 1.25$ sec.}
	\label{tip_gait_shape02}
	\end{subfigure}
	\hfill
  \begin{subfigure}{0.65\columnwidth}
  \centering
  \resizebox{1.0\columnwidth}{!}{\input{tip_gait_shape_3.tex}} 
	\caption{$t = 4.3$ sec.}
	\label{tip_gait_shape03}
	\end{subfigure}
	\caption{Sequence of Turn-in-Place gait shapes and component curves, over the course of one gait cycle ($T = 5$ sec.): reference body curve $\abProject(s, t)$ (dashed brown); planar body curve $\gSpace(s)$ (green+purple); body-ground contact segments $\mathcal{S}_{\rm cont}(t)$ (purple).} 
	\label{tip_gait_shape}
\end{figure*}

Recognizing the need for locomotion that goes beyond primarily
translational movement for snake-like robots has led to the design of
gaits with large angular and small translational displacements, 
i.e., turning in-place.  Such gaits complement the gaits traditionally
used for snake-like robot locomotion that achieve large translational
displacements to follow reference paths and reach goal positions many
body lengths distant. Two examples being the Omega Turn
as engineered from biologically-inspired observations of worms
\cite{WaEtAl_IROS[2020]}, and `frequency turning' as discovered
through a parameter sweep of a parametrized gait \cite{DaEtAl_ICRA[2015]}. 
Gait shapes associated with both are formulated as traveling waves and
vary in their use of vertical lift to manage body-ground contact. 
Here, we design a Turn-in-Place gait from basic physical principles. A
standing wave gait shape is coupled with a coordinated body-ground
contact profile to generate periodic stepping motion, whereby each step
rotates the snake in-place.  Modeling of the gait dynamics employs a
shape-centric modeling framework \cite{ChVe_RAS[2020]} to confirm
Turn-in-Place movement.

The full 3-D gait shape is represented by a continuous-body curve,
$\bCurvePos$, defined relative to a reference body curve, $\abCurvePos$,
and rigidly attached body frame.  Analysis of the gait's locomotive
properties focuses on the gait shape projection of $\bCurvePos$ onto the
plane of locomotion. This projected shape is denoted by $\gSpacePos$.
The component of $\bCurvePos$ describing vertical lifting of the body
from the ground will be described by a body-ground contact profile,
defined along $\gSpace(s, t)$.  Under this model, design of the
Turn-in-Place gait kinematics reduces to engineering a time-cyclical
body shape, $\gSpace(s, t)$, and a coordinated body-ground contact
profile $\mathcal{S}_{\rm cont}(t)$. 

\subsection{Gait Shape}
The Turn-in-Place gait shape description in the locomotion modeling plane, $\gSpacePos$, is a sinusoidal
standing wave; see Figure \ref{tip_gait_shape}. Its planar shape is
defined with respect to the reference body curve's projection onto the
locomotion plane, $\abProject(s, t)$. For this gait, $\abProject(s, t)$
is a straight line parametrized with respect to arc length, $s$. The
body frame $g \in SE(2)$ is rigidly attached to the reference body at
the center of the reference body where $s = 0$. The length of the
reference body is $L$ such that $s \in \left[ -\frac{L}{2}, \frac{L}{2}
\right]$. The Turn-in-Place gait standing wave is 
\begin{equation} \eqlabel{gait_shape}
  \gSpacePos = 
  \mymatrix{s \\ A(t) \sin \left( 2 \pi \cdot \frac{s}{\lambda(t)} \right)}
\end{equation}
where
\begin{equation} \eqlabel{gait_shape_Ah}
  A(t) = \bar{A} \sin(\pi f t) 
\end{equation}
for peak amplitude $\bar{A}$, frequency $f$ and corresponding gait
period  $T = \frac{1}{f}$.
The wavelength, $\lambda(t)$, is a function of $A(t)$ and the total
fixed curve length $\bar{L}$ (i.e., robot body length). 
It is computed by the root-finding operation, $\lambda(t) = {\rm
root}_{\gamma} \left( {\rm Arclen}\left( \gSpace(s, t; A(t), \gamma)
\right) - \bar{L} \right)$.  The gait is designed to be a single
sinusoidal standing wave, $L(t) = \lambda(t)$, periodic in time.

The robot begins each gait cycle in a straight-body configuration, illustrated in Figure \ref{tip_gait_shape01}, where $\gSpace(s, t)$ is denoted by the union of purple and green curve segments. Over the course of a half gait period, the shape transitions to a sine wave comprising a single wavelength, $\lambda$ (Fig. \ref{tip_gait_shape02}). During the second half of the gait cycle, this shape transitions back to a straight-body configuration (Fig. \ref{tip_gait_shape03}).

The rigid reference frame and the gait shape define a set of local
frames $\gshape(s, t) \in SE(2)$ residing along the body curve $\gSpace(s, t)$ 
whose local $x$-axis is tangent to the body curve. The frame decomposes
into a rotation, $R(s, t) \in SO(2)$, and a translation,
$\gSpace(s, t) \in E(2)$, specified relative to the rigid body frame, $g$. 

A vertical body curve component, $h(s, t)$, dictates which segments of the
body are in ground contact and when. It is constructed such that the segments in ground contact 
follow a desired contact schedule $\mathcal{S}_{\rm cont}(t)$, in \eqref{tip_contact}, for this gait.
The full 3-D gait shape is described as, $\bCurvePos = \gSpacePos + h(s, t)$.

\begin{figure*}[t]
  \centering
  \normalsize
  \begin{equation} \eqlabel{tip_contact}
    \mathcal{S}_{\rm cont}(t) = 
      \begin{cases}
        \left\{ s \in \left[-\frac{\ell_{\rm cont}}{2}, \frac{\ell_{\rm cont}}{2}\right] \cup \left[\frac{L}{2}-\frac{\ell_{\rm cont}}{2}, \frac{L}{2}+\frac{\ell_{\rm cont}}{2}\right] \right\} & t \in [0, \frac{T}{2}) \\
        \left\{ s \in \left[ -\frac{L}{2}, \frac{L}{2} \right] \right\} & t \in [\frac{T}{2}, T) 
      \end{cases}, 
      \text{where } \ell_{\rm cont}(t) = \max \left( \frac{L}{4} \left( 1+\cos\left(4 \pi f t\right) \right), \bar{\ell}_{\rm cont} \right)
    \end{equation}
  \hrulefill
  \vspace*{-0.8em}
\end{figure*}

\subsection{Robot-Environment Contact}
Vertical shape deformations of the snake-robot lead to desired
robot-ground interaction forces. 
Here, the time-periodic body-ground contact
pattern is engineered to promote in-plane rotation through a stepping
pattern coupled to asymmetrical contact velocities.  In reference to
Figure \ref{tip_gait_shape}, body-ground contact segments are
illustrated as purple curves (green means no contact). 
The set of points along the body with ground contact is represented by 
\eqref{tip_contact}, where $\ell_{\rm cont}(t)$ denotes the
cyclically-varying length of each contact segment. Both contact segments
maintain a minimum length $\bar{\ell}_{\rm cont}$, and are centered
about two points along the body, at $s = 0, \frac{L}{2}$. A ground
contact segment extending beyond the head of the body, at $s =
\frac{L}{2}$, wraps-around to continue from the tail end, at $s =
-\frac{L}{2}$. 

Time-coordination between the gait shape $\gSpace(s,  t)$ in \eqref{gait_shape}
and 
body-ground contact $\mathcal{S}_{\rm cont}(t)$ in \eqref{tip_contact}
results in a robot motion pattern that appears like `rotational stepping'. 
The robot begins with a straight-body at the start of the gait. 
Over $t \in [0, \frac{T}{2})$ it picks up interior segments of the body
as $\gSpace(s, t)$ evolves following a standing wave configuration with
peak amplitude, $\bar{A}$. Over $t \in [\frac{T}{2}, T)$ the entire body
achieves ground contact as the gait shape recedes back to a
straight-line configuration. This motion pattern produces a net change
in orientation over a single gait cycle, with little translational
displacement.  We presume the Turn-in-Place gait evolves in a
quasi-static manner; contact segments, $\mathcal{S}_{\rm cont}(t)$, provide mechanical stability for the robot throughout
the gait cycle.

%% file: tip_gait_shape_2.tex
%
%
\definecolor{mycolor1}{rgb}{0.85000,0.32500,0.09800}%
\definecolor{mycolor2}{rgb}{0.70000,0.00000,0.70000}%
\begin{tikzpicture}

\begin{axis}[%
width=6.000in,
height=3.000in,
at={(1.213in,2.772in)},
scale only axis,
xmin=-425,
xmax=425,
ymin=-100,
ymax=100,
axis background/.style={fill=white},
xtick=\empty,
xticklabels=\empty,
ytick=\empty,
yticklabels=\empty,
hide x axis,
hide y axis,
]
\addplot [color=mycolor1, dashed, line width=2.0pt, forget plot]
  table[row sep=crcr]{%
-390	0\\
-388.045112781955	0\\
-386.09022556391	0\\
-384.135338345865	0\\
-382.18045112782	0\\
-380.225563909774	0\\
-378.270676691729	0\\
-376.315789473684	0\\
-374.360902255639	0\\
-372.406015037594	0\\
-370.451127819549	0\\
-368.496240601504	0\\
-366.541353383459	0\\
-364.586466165414	0\\
-362.631578947368	0\\
-360.676691729323	0\\
-358.721804511278	0\\
-356.766917293233	0\\
-354.812030075188	0\\
-352.857142857143	0\\
-350.902255639098	0\\
-348.947368421053	0\\
-346.992481203008	0\\
-345.037593984962	0\\
-343.082706766917	0\\
-341.127819548872	0\\
-339.172932330827	0\\
-337.218045112782	0\\
-335.263157894737	0\\
-333.308270676692	0\\
-331.353383458647	0\\
-329.398496240602	0\\
-327.443609022556	0\\
-325.488721804511	0\\
-323.533834586466	0\\
-321.578947368421	0\\
-319.624060150376	0\\
-317.669172932331	0\\
-315.714285714286	0\\
-313.759398496241	0\\
-311.804511278195	0\\
-309.84962406015	0\\
-307.894736842105	0\\
-305.93984962406	0\\
-303.984962406015	0\\
-302.03007518797	0\\
-300.075187969925	0\\
-298.12030075188	0\\
-296.165413533835	0\\
-294.210526315789	0\\
-292.255639097744	0\\
-290.300751879699	0\\
-288.345864661654	0\\
-286.390977443609	0\\
-284.436090225564	0\\
-282.481203007519	0\\
-280.526315789474	0\\
-278.571428571429	0\\
-276.616541353383	0\\
-274.661654135338	0\\
-272.706766917293	0\\
-270.751879699248	0\\
-268.796992481203	0\\
-266.842105263158	0\\
-264.887218045113	0\\
-262.932330827068	0\\
-260.977443609023	0\\
-259.022556390977	0\\
-257.067669172932	0\\
-255.112781954887	0\\
-253.157894736842	0\\
-251.203007518797	0\\
-249.248120300752	0\\
-247.293233082707	0\\
-245.338345864662	0\\
-243.383458646617	0\\
-241.428571428571	0\\
-239.473684210526	0\\
-237.518796992481	0\\
-235.563909774436	0\\
-233.609022556391	0\\
-231.654135338346	0\\
-229.699248120301	0\\
-227.744360902256	0\\
-225.789473684211	0\\
-223.834586466165	0\\
-221.87969924812	0\\
-219.924812030075	0\\
-217.96992481203	0\\
-216.015037593985	0\\
-214.06015037594	0\\
-212.105263157895	0\\
-210.15037593985	0\\
-208.195488721805	0\\
-206.240601503759	0\\
-204.285714285714	0\\
-202.330827067669	0\\
-200.375939849624	0\\
-198.421052631579	0\\
-196.466165413534	0\\
-194.511278195489	0\\
-192.556390977444	0\\
-190.601503759398	0\\
-188.646616541353	0\\
-186.691729323308	0\\
-184.736842105263	0\\
-182.781954887218	0\\
-180.827067669173	0\\
-178.872180451128	0\\
-176.917293233083	0\\
-174.962406015038	0\\
-173.007518796992	0\\
-171.052631578947	0\\
-169.097744360902	0\\
-167.142857142857	0\\
-165.187969924812	0\\
-163.233082706767	0\\
-161.278195488722	0\\
-159.323308270677	0\\
-157.368421052632	0\\
-155.413533834586	0\\
-153.458646616541	0\\
-151.503759398496	0\\
-149.548872180451	0\\
-147.593984962406	0\\
-145.639097744361	0\\
-143.684210526316	0\\
-141.729323308271	0\\
-139.774436090226	0\\
-137.81954887218	0\\
-135.864661654135	0\\
-133.90977443609	0\\
-131.954887218045	0\\
-130	0\\
-128.045112781955	0\\
-126.09022556391	0\\
-124.135338345865	0\\
-122.18045112782	0\\
-120.225563909774	0\\
-118.270676691729	0\\
-116.315789473684	0\\
-114.360902255639	0\\
-112.406015037594	0\\
-110.451127819549	0\\
-108.496240601504	0\\
-106.541353383459	0\\
-104.586466165414	0\\
-102.631578947368	0\\
-100.676691729323	0\\
-98.7218045112782	0\\
-96.7669172932331	0\\
-94.812030075188	0\\
-92.8571428571428	0\\
-90.9022556390977	0\\
-88.9473684210526	0\\
-86.9924812030075	0\\
-85.0375939849624	0\\
-83.0827067669173	0\\
-81.1278195488722	0\\
-79.1729323308271	0\\
-77.218045112782	0\\
-75.2631578947368	0\\
-73.3082706766917	0\\
-71.3533834586466	0\\
-69.3984962406015	0\\
-67.4436090225564	0\\
-65.4887218045113	0\\
-63.5338345864662	0\\
-61.578947368421	0\\
-59.6240601503759	0\\
-57.6691729323308	0\\
-55.7142857142857	0\\
-53.7593984962406	0\\
-51.8045112781955	0\\
-49.8496240601504	0\\
-47.8947368421053	0\\
-45.9398496240601	0\\
-43.984962406015	0\\
-42.0300751879699	0\\
-40.0751879699248	0\\
-38.1203007518797	0\\
-36.1654135338346	0\\
-34.2105263157895	0\\
-32.2556390977444	0\\
-30.3007518796992	0\\
-28.3458646616541	0\\
-26.390977443609	0\\
-24.4360902255639	0\\
-22.4812030075188	0\\
-20.5263157894737	0\\
-18.5714285714286	0\\
-16.6165413533835	0\\
-14.6616541353383	0\\
-12.7067669172932	0\\
-10.7518796992481	0\\
-8.79699248120301	0\\
-6.84210526315789	0\\
-4.88721804511278	0\\
-2.93233082706767	0\\
-0.977443609022556	0\\
0.977443609022556	0\\
2.93233082706767	0\\
4.88721804511278	0\\
6.84210526315789	0\\
8.79699248120301	0\\
10.7518796992481	0\\
12.7067669172932	0\\
14.6616541353383	0\\
16.6165413533835	0\\
18.5714285714286	0\\
20.5263157894737	0\\
22.4812030075188	0\\
24.4360902255639	0\\
26.390977443609	0\\
28.3458646616541	0\\
30.3007518796992	0\\
32.2556390977444	0\\
34.2105263157895	0\\
36.1654135338346	0\\
38.1203007518797	0\\
40.0751879699248	0\\
42.0300751879699	0\\
43.984962406015	0\\
45.9398496240601	0\\
47.8947368421053	0\\
49.8496240601504	0\\
51.8045112781955	0\\
53.7593984962406	0\\
55.7142857142857	0\\
57.6691729323308	0\\
59.6240601503759	0\\
61.578947368421	0\\
63.5338345864662	0\\
65.4887218045113	0\\
67.4436090225564	0\\
69.3984962406015	0\\
71.3533834586466	0\\
73.3082706766917	0\\
75.2631578947368	0\\
77.218045112782	0\\
79.1729323308271	0\\
81.1278195488722	0\\
83.0827067669173	0\\
85.0375939849624	0\\
86.9924812030075	0\\
88.9473684210526	0\\
90.9022556390977	0\\
92.8571428571428	0\\
94.812030075188	0\\
96.7669172932331	0\\
98.7218045112782	0\\
100.676691729323	0\\
102.631578947368	0\\
104.586466165414	0\\
106.541353383459	0\\
108.496240601504	0\\
110.451127819549	0\\
112.406015037594	0\\
114.360902255639	0\\
116.315789473684	0\\
118.270676691729	0\\
120.225563909774	0\\
122.18045112782	0\\
124.135338345865	0\\
126.09022556391	0\\
128.045112781955	0\\
130	0\\
131.954887218045	0\\
133.90977443609	0\\
135.864661654135	0\\
137.81954887218	0\\
139.774436090226	0\\
141.729323308271	0\\
143.684210526316	0\\
145.639097744361	0\\
147.593984962406	0\\
149.548872180451	0\\
151.503759398496	0\\
153.458646616541	0\\
155.413533834586	0\\
157.368421052632	0\\
159.323308270677	0\\
161.278195488722	0\\
163.233082706767	0\\
165.187969924812	0\\
167.142857142857	0\\
169.097744360902	0\\
171.052631578947	0\\
173.007518796992	0\\
174.962406015038	0\\
176.917293233083	0\\
178.872180451128	0\\
180.827067669173	0\\
182.781954887218	0\\
184.736842105263	0\\
186.691729323308	0\\
188.646616541353	0\\
190.601503759398	0\\
192.556390977444	0\\
194.511278195489	0\\
196.466165413534	0\\
198.421052631579	0\\
200.375939849624	0\\
202.330827067669	0\\
204.285714285714	0\\
206.240601503759	0\\
208.195488721805	0\\
210.15037593985	0\\
212.105263157895	0\\
214.06015037594	0\\
216.015037593985	0\\
217.96992481203	0\\
219.924812030075	0\\
221.87969924812	0\\
223.834586466165	0\\
225.789473684211	0\\
227.744360902256	0\\
229.699248120301	0\\
231.654135338346	0\\
233.609022556391	0\\
235.563909774436	0\\
237.518796992481	0\\
239.473684210526	0\\
241.428571428571	0\\
243.383458646617	0\\
245.338345864662	0\\
247.293233082707	0\\
249.248120300752	0\\
251.203007518797	0\\
253.157894736842	0\\
255.112781954887	0\\
257.067669172932	0\\
259.022556390977	0\\
260.977443609023	0\\
262.932330827068	0\\
264.887218045113	0\\
266.842105263158	0\\
268.796992481203	0\\
270.751879699248	0\\
272.706766917293	0\\
274.661654135338	0\\
276.616541353383	0\\
278.571428571429	0\\
280.526315789474	0\\
282.481203007519	0\\
284.436090225564	0\\
286.390977443609	0\\
288.345864661654	0\\
290.300751879699	0\\
292.255639097744	0\\
294.210526315789	0\\
296.165413533835	0\\
298.12030075188	0\\
300.075187969925	0\\
302.03007518797	0\\
303.984962406015	0\\
305.93984962406	0\\
307.894736842105	0\\
309.84962406015	0\\
311.804511278195	0\\
313.759398496241	0\\
315.714285714286	0\\
317.669172932331	0\\
319.624060150376	0\\
321.578947368421	0\\
323.533834586466	0\\
325.488721804511	0\\
327.443609022556	0\\
329.398496240602	0\\
331.353383458647	0\\
333.308270676692	0\\
335.263157894737	0\\
337.218045112782	0\\
339.172932330827	0\\
341.127819548872	0\\
343.082706766917	0\\
345.037593984962	0\\
346.992481203008	0\\
348.947368421053	0\\
350.902255639098	0\\
352.857142857143	0\\
354.812030075188	0\\
356.766917293233	0\\
358.721804511278	0\\
360.676691729323	0\\
362.631578947368	0\\
364.586466165414	0\\
366.541353383459	0\\
368.496240601504	0\\
370.451127819549	0\\
372.406015037594	0\\
374.360902255639	0\\
376.315789473684	0\\
378.270676691729	0\\
380.225563909774	0\\
382.18045112782	0\\
384.135338345865	0\\
386.09022556391	0\\
388.045112781955	0\\
390	0\\
};
\addplot [color=black!30!green, line width=2.0pt, forget plot]
  table[row sep=crcr]{%
-366.878655766713	-7.34759071599491e-15\\
-365.03966501099	-0.944763547559934\\
-363.200674255268	-1.8892928189588\\
-361.361683499545	-2.83335359612972\\
-359.522692743822	-3.77671177717993\\
-357.683701988099	-4.71913343444177\\
-355.844711232376	-5.66038487248051\\
-354.005720476653	-6.60023268604451\\
-352.16672972093	-7.53844381794361\\
-350.327738965207	-8.47478561684108\\
-348.488748209485	-9.40902589494483\\
-346.649757453762	-10.340932985584\\
-344.810766698039	-11.2702758006558\\
-342.971775942316	-12.1968238879296\\
-341.132785186593	-13.1203474881926\\
-339.29379443087	-14.0406175922241\\
-337.454803675147	-14.9574059975837\\
-335.615812919424	-15.8704853651996\\
-333.776822163702	-16.7796292757421\\
-331.937831407979	-17.68461228577\\
-330.098840652256	-18.5852099836347\\
-328.259849896533	-19.4811990451276\\
-326.42085914081	-20.3723572888595\\
-324.581868385087	-21.2584637313548\\
-322.742877629364	-22.13929864185\\
-320.903886873642	-23.0146435967808\\
-319.064896117919	-23.8842815339457\\
-317.225905362196	-24.7479968063311\\
-315.386914606473	-25.6055752355865\\
-313.54792385075	-26.4568041651348\\
-311.708933095027	-27.3014725129054\\
-309.869942339304	-28.1393708236771\\
-308.030951583581	-28.9702913210171\\
-306.191960827859	-29.7940279588044\\
-304.352970072136	-30.6103764723235\\
-302.513979316413	-31.4191344289168\\
-300.67498856069	-32.220101278182\\
-298.835997804967	-33.0130784017041\\
-296.997007049244	-33.7978691623066\\
-295.158016293521	-34.5742789528128\\
-293.319025537798	-35.3421152443031\\
-291.480034782076	-36.1011876338568\\
-289.641044026353	-36.8513078917674\\
-287.80205327063	-37.592290008218\\
-285.963062514907	-38.3239502394072\\
-284.124071759184	-39.0461071531126\\
-282.285081003461	-39.7585816736809\\
-280.446090247738	-40.461197126434\\
-278.607099492015	-41.1537792814798\\
-276.768108736293	-41.8361563969165\\
-274.92911798057	-42.5081592614197\\
-273.090127224847	-43.1696212362028\\
-271.251136469124	-43.8203782963383\\
-269.412145713401	-44.4602690714325\\
-267.573154957678	-45.0891348856401\\
-265.734164201955	-45.7068197970125\\
-263.895173446232	-46.3131706361664\\
-262.05618269051	-46.9080370442664\\
-260.217191934787	-47.4912715103094\\
-258.378201179064	-48.062729407704\\
-256.539210423341	-48.6222690301334\\
-254.700219667618	-49.1697516266954\\
-252.861228911895	-49.7050414363083\\
-251.022238156172	-50.2280057213768\\
-249.183247400449	-50.7385148007066\\
-247.344256644727	-51.2364420816622\\
-245.505265889004	-51.7216640915588\\
-243.666275133281	-52.1940605082797\\
-241.827284377558	-52.6535141901132\\
-239.988293621835	-53.0999112048009\\
-238.149302866112	-53.5331408577894\\
-236.310312110389	-53.9530957196798\\
-234.471321354666	-54.3596716528674\\
-232.632330598944	-54.752767837365\\
-230.793339843221	-55.1322867958035\\
-228.954349087498	-55.4981344176036\\
-227.115358331775	-55.8502199823131\\
-225.276367576052	-56.1884561821028\\
-223.437376820329	-56.5127591434164\\
-221.598386064606	-56.8230484477695\\
-219.759395308883	-57.1192471516905\\
-217.920404553161	-57.4012818058011\\
-216.081413797438	-57.6690824730295\\
-214.242423041715	-57.922582745953\\
-212.403432285992	-58.1617197632653\\
-210.564441530269	-58.3864342253643\\
-208.725450774546	-58.596670409057\\
-206.886460018823	-58.7923761813772\\
-205.0474692631	-58.9735030125133\\
-203.208478507378	-59.1400059878422\\
-201.369487751655	-59.291843819067\\
-199.530496995932	-59.4289788544553\\
-197.691506240209	-59.551377088176\\
-195.852515484486	-59.6590081687318\\
-194.013524728763	-59.7518454064853\\
-192.17453397304	-59.8298657802779\\
-190.335543217317	-59.8930499431377\\
-188.496552461595	-59.9413822270777\\
-186.657561705872	-59.9748506469804\\
-184.818570950149	-59.9934469035705\\
-182.979580194426	-59.9971663854723\\
-181.140589438703	-59.9860081703535\\
-179.30159868298	-59.9599750251539\\
-177.462607927257	-59.919073405399\\
-175.623617171534	-59.8633134535998\\
-173.784626415812	-59.7927089967369\\
-171.945635660089	-59.7072775428325\\
-170.106644904366	-59.6070402766087\\
-168.267654148643	-59.4920220542338\\
-166.42866339292	-59.3622513971592\\
-164.589672637197	-59.2177604850467\\
-162.750681881474	-59.0585851477886\\
-160.911691125751	-58.8847648566231\\
-159.072700370029	-58.6963427143462\\
-157.233709614306	-58.4933654446236\\
-155.394718858583	-58.2758833804043\\
-153.55572810286	-58.0439504514398\\
-151.716737347137	-57.7976241709101\\
-149.877746591414	-57.5369656211629\\
-148.038755835691	-57.2620394385663\\
-146.199765079968	-56.9729137974806\\
-144.360774324246	-56.6696603933534\\
-142.521783568523	-56.3523544249405\\
-140.6827928128	-56.0210745756589\\
-138.843802057077	-55.6759029940752\\
-137.004811301354	-55.3169252735355\\
-135.165820545631	-54.9442304309395\\
-133.326829789908	-54.557910884668\\
-131.487839034185	-54.1580624316643\\
-129.648848278463	-53.7447842236802\\
-127.80985752274	-53.3181787426882\\
-125.970866767017	-52.8783517754692\\
-124.131876011294	-52.42541238738\\
-122.292885255571	-51.9594728953082\\
-120.453894499848	-51.4806488398201\\
-118.614903744125	-50.9890589565104\\
-116.775912988402	-50.4848251465585\\
-114.93692223268	-49.9680724465003\\
-113.097931476957	-49.4389289972226\\
-111.258940721234	-48.8975260121875\\
-109.419949965511	-48.343997744895\\
-107.580959209788	-47.7784814555918\\
-105.741968454065	-47.2011173772342\\
-103.902977698342	-46.612048680714\\
-102.063986942619	-46.0114214393566\\
-100.224996186897	-45.3993845926977\\
-98.3860054311737	-44.7760899095511\\
-96.5470146754509	-44.1416919503737\\
-94.708023919728	-43.4963480289389\\
-92.8690331640051	-42.8402181733264\\
-91.0300424082823	-42.1734650862402\\
-89.1910516525594	-41.4962541046623\\
-87.3520608968365	-40.8087531588536\\
-85.5130701411136	-40.1111327307115\\
-83.6740793853908	-39.4035658114954\\
-81.8350886296679	-38.6862278589291\\
-79.996097873945	-37.9592967536919\\
-78.1571071182221	-37.2229527553092\\
-76.3181163624993	-36.477378457453\\
-74.4791256067764	-35.722758742663\\
-72.6401348510535	-34.9592807365015\\
-70.8011440953306	-34.1871337611504\\
-68.9621533396078	-33.406509288465\\
-67.1231625838849	-32.6176008924936\\
-65.284171828162	-31.8206042014768\\
-63.4451810724391	-31.0157168493365\\
-61.6061903167163	-30.2031384266679\\
-59.7671995609934	-29.383070431247\\
-57.9282088052705	-28.5557162180636\\
-56.0892180495476	-27.7212809488956\\
-54.2502272938248	-26.8799715414336\\
-52.4112365381019	-26.0319966179713\\
-50.572245782379	-25.1775664536725\\
-48.7332550266562	-24.3168929244287\\
-46.8942642709333	-23.4501894543192\\
-45.0552735152104	-22.5776709626879\\
-43.2162827594875	-21.699553810849\\
-41.3772920037647	-20.816055748435\\
-39.5383012480418	-19.9273958594007\\
-37.6993104923189	-19.0337945076963\\
-35.860319736596	-18.135473282623\\
-34.0213289808732	-17.2326549438844\\
-32.1823382251503	-16.3255633663488\\
-30.3433474694274	-15.4144234845333\\
-28.5043567137045	-14.4994612368268\\
-26.6653659579817	-13.5809035094628\\
-24.8263752022588	-12.6589780802581\\
-22.9873844465359	-11.7339135621294\\
-21.148393690813	-10.8059393464039\\
-19.3094029350902	-9.87528554593634\\
-17.4704121793673	-8.94218293804674\\
-15.6314214236444	-8.00686290729419\\
-13.7924306679216	-7.0695573880995\\
-11.9534399121987	-6.13049880723162\\
-10.1144491564758	-5.18992002617207\\
-8.27545840075293	-4.24805428337145\\
-6.43646764503006	-3.30513513641256\\
-4.59747688930718	-2.36139640409433\\
-2.75848613358431	-1.41707210845098\\
-0.919495377861437	-0.472396416720862\\
0.919495377861437	0.472396416720862\\
2.75848613358431	1.41707210845098\\
4.59747688930718	2.36139640409433\\
6.43646764503006	3.30513513641256\\
8.27545840075293	4.24805428337145\\
10.1144491564758	5.18992002617207\\
11.9534399121987	6.13049880723162\\
13.7924306679216	7.0695573880995\\
15.6314214236444	8.00686290729419\\
17.4704121793673	8.94218293804674\\
19.3094029350902	9.87528554593634\\
21.148393690813	10.8059393464039\\
22.9873844465359	11.7339135621294\\
24.8263752022588	12.6589780802581\\
26.6653659579817	13.5809035094628\\
28.5043567137045	14.4994612368268\\
30.3433474694274	15.4144234845333\\
32.1823382251503	16.3255633663488\\
34.0213289808732	17.2326549438844\\
35.860319736596	18.135473282623\\
37.6993104923189	19.0337945076963\\
39.5383012480418	19.9273958594007\\
41.3772920037647	20.816055748435\\
43.2162827594875	21.699553810849\\
45.0552735152104	22.5776709626879\\
46.8942642709333	23.4501894543192\\
48.7332550266562	24.3168929244287\\
50.572245782379	25.1775664536725\\
52.4112365381019	26.0319966179713\\
54.2502272938248	26.8799715414336\\
56.0892180495476	27.7212809488956\\
57.9282088052705	28.5557162180636\\
59.7671995609934	29.383070431247\\
61.6061903167163	30.2031384266679\\
63.4451810724391	31.0157168493365\\
65.284171828162	31.8206042014768\\
67.1231625838849	32.6176008924936\\
68.9621533396078	33.406509288465\\
70.8011440953306	34.1871337611504\\
72.6401348510535	34.9592807365015\\
74.4791256067764	35.722758742663\\
76.3181163624993	36.477378457453\\
78.1571071182221	37.2229527553092\\
79.996097873945	37.9592967536919\\
81.8350886296679	38.6862278589291\\
83.6740793853908	39.4035658114954\\
85.5130701411136	40.1111327307115\\
87.3520608968365	40.8087531588536\\
89.1910516525594	41.4962541046623\\
91.0300424082823	42.1734650862402\\
92.8690331640051	42.8402181733264\\
94.708023919728	43.4963480289389\\
96.5470146754509	44.1416919503737\\
98.3860054311737	44.7760899095511\\
100.224996186897	45.3993845926977\\
102.063986942619	46.0114214393566\\
103.902977698342	46.612048680714\\
105.741968454065	47.2011173772342\\
107.580959209788	47.7784814555918\\
109.419949965511	48.343997744895\\
111.258940721234	48.8975260121875\\
113.097931476957	49.4389289972226\\
114.93692223268	49.9680724465003\\
116.775912988402	50.4848251465585\\
118.614903744125	50.9890589565104\\
120.453894499848	51.4806488398201\\
122.292885255571	51.9594728953082\\
124.131876011294	52.42541238738\\
125.970866767017	52.8783517754692\\
127.80985752274	53.3181787426882\\
129.648848278463	53.7447842236802\\
131.487839034185	54.1580624316643\\
133.326829789908	54.557910884668\\
135.165820545631	54.9442304309395\\
137.004811301354	55.3169252735355\\
138.843802057077	55.6759029940752\\
140.6827928128	56.0210745756589\\
142.521783568523	56.3523544249405\\
144.360774324246	56.6696603933534\\
146.199765079968	56.9729137974806\\
148.038755835691	57.2620394385663\\
149.877746591414	57.5369656211629\\
151.716737347137	57.7976241709101\\
153.55572810286	58.0439504514398\\
155.394718858583	58.2758833804043\\
157.233709614306	58.4933654446236\\
159.072700370029	58.6963427143462\\
160.911691125751	58.8847648566231\\
162.750681881474	59.0585851477886\\
164.589672637197	59.2177604850467\\
166.42866339292	59.3622513971592\\
168.267654148643	59.4920220542338\\
170.106644904366	59.6070402766087\\
171.945635660089	59.7072775428325\\
173.784626415812	59.7927089967369\\
175.623617171534	59.8633134535998\\
177.462607927257	59.919073405399\\
179.30159868298	59.9599750251539\\
181.140589438703	59.9860081703535\\
182.979580194426	59.9971663854723\\
184.818570950149	59.9934469035705\\
186.657561705872	59.9748506469804\\
188.496552461595	59.9413822270777\\
190.335543217317	59.8930499431377\\
192.17453397304	59.8298657802779\\
194.013524728763	59.7518454064853\\
195.852515484486	59.6590081687318\\
197.691506240209	59.551377088176\\
199.530496995932	59.4289788544553\\
201.369487751655	59.291843819067\\
203.208478507378	59.1400059878422\\
205.0474692631	58.9735030125133\\
206.886460018823	58.7923761813772\\
208.725450774546	58.596670409057\\
210.564441530269	58.3864342253643\\
212.403432285992	58.1617197632653\\
214.242423041715	57.922582745953\\
216.081413797438	57.6690824730295\\
217.920404553161	57.4012818058011\\
219.759395308883	57.1192471516905\\
221.598386064606	56.8230484477695\\
223.437376820329	56.5127591434164\\
225.276367576052	56.1884561821028\\
227.115358331775	55.8502199823131\\
228.954349087498	55.4981344176036\\
230.793339843221	55.1322867958035\\
232.632330598944	54.752767837365\\
234.471321354666	54.3596716528674\\
236.310312110389	53.9530957196798\\
238.149302866112	53.5331408577894\\
239.988293621835	53.0999112048009\\
241.827284377558	52.6535141901132\\
243.666275133281	52.1940605082797\\
245.505265889004	51.7216640915588\\
247.344256644727	51.2364420816622\\
249.183247400449	50.7385148007066\\
251.022238156172	50.2280057213768\\
252.861228911895	49.7050414363083\\
254.700219667618	49.1697516266954\\
256.539210423341	48.6222690301334\\
258.378201179064	48.062729407704\\
260.217191934787	47.4912715103094\\
262.05618269051	46.9080370442664\\
263.895173446232	46.3131706361664\\
265.734164201955	45.7068197970125\\
267.573154957678	45.0891348856401\\
269.412145713401	44.4602690714325\\
271.251136469124	43.8203782963383\\
273.090127224847	43.1696212362028\\
274.92911798057	42.5081592614197\\
276.768108736293	41.8361563969165\\
278.607099492015	41.1537792814798\\
280.446090247738	40.461197126434\\
282.285081003461	39.7585816736809\\
284.124071759184	39.0461071531126\\
285.963062514907	38.3239502394072\\
287.80205327063	37.592290008218\\
289.641044026353	36.8513078917674\\
291.480034782076	36.1011876338568\\
293.319025537798	35.3421152443031\\
295.158016293521	34.5742789528128\\
296.997007049244	33.7978691623066\\
298.835997804967	33.0130784017041\\
300.67498856069	32.220101278182\\
302.513979316413	31.4191344289168\\
304.352970072136	30.6103764723235\\
306.191960827859	29.7940279588044\\
308.030951583581	28.9702913210171\\
309.869942339304	28.1393708236771\\
311.708933095027	27.3014725129054\\
313.54792385075	26.4568041651348\\
315.386914606473	25.6055752355865\\
317.225905362196	24.7479968063311\\
319.064896117919	23.8842815339457\\
320.903886873642	23.0146435967808\\
322.742877629364	22.13929864185\\
324.581868385087	21.2584637313548\\
326.42085914081	20.3723572888595\\
328.259849896533	19.4811990451276\\
330.098840652256	18.5852099836347\\
331.937831407979	17.68461228577\\
333.776822163702	16.7796292757421\\
335.615812919424	15.8704853651996\\
337.454803675147	14.9574059975837\\
339.29379443087	14.0406175922241\\
341.132785186593	13.1203474881926\\
342.971775942316	12.1968238879296\\
344.810766698039	11.2702758006558\\
346.649757453762	10.340932985584\\
348.488748209485	9.40902589494483\\
350.327738965207	8.47478561684108\\
352.16672972093	7.53844381794361\\
354.005720476653	6.60023268604451\\
355.844711232376	5.66038487248051\\
357.683701988099	4.71913343444177\\
359.522692743822	3.77671177717993\\
361.361683499545	2.83335359612972\\
363.200674255268	1.8892928189588\\
365.03966501099	0.944763547559934\\
366.878655766713	7.34759071599491e-15\\
};
\addplot [color=mycolor2, line width=3.0pt, forget plot]
  table[row sep=crcr]{%
-18.3439327883357	-9.38569736018172\\
-17.3248254112059	-8.86821455548447\\
-16.3057180340762	-8.35005640191412\\
-15.2866106569464	-7.83126235922021\\
-14.2675032798166	-7.31187193557768\\
-13.2483959026869	-6.7919246845782\\
-12.2292885255571	-6.27146020221798\\
-11.2101811484274	-5.75051812388241\\
-10.1910737712976	-5.22913812132767\\
-9.17196639416784	-4.70735989965957\\
-8.15285901703808	-4.18522319430982\\
-7.13375163990832	-3.66276776801007\\
-6.11464426277856	-3.14003340776381\\
-5.0955368856488	-2.61705992181641\\
-4.07642950851904	-2.0938871366236\\
-3.05732213138928	-1.57055489381853\\
-2.03821475425952	-1.04710304717764\\
-1.01910737712976	-0.52357145958569\\
0	0\\
1.01910737712976	0.52357145958569\\
2.03821475425952	1.04710304717764\\
3.05732213138928	1.57055489381853\\
4.07642950851904	2.0938871366236\\
5.0955368856488	2.61705992181641\\
6.11464426277856	3.14003340776381\\
7.13375163990832	3.66276776801007\\
8.15285901703808	4.18522319430982\\
9.17196639416784	4.70735989965957\\
10.1910737712976	5.22913812132767\\
11.2101811484274	5.75051812388241\\
12.2292885255571	6.27146020221798\\
13.2483959026869	6.7919246845782\\
14.2675032798166	7.31187193557768\\
15.2866106569464	7.83126235922021\\
16.3057180340762	8.35005640191412\\
17.3248254112059	8.86821455548447\\
18.3439327883357	9.38569736018172\\
};
\addplot [color=mycolor2, line width=3.0pt, forget plot]
  table[row sep=crcr]{%
-366.878655766713	-7.34759071599491e-15\\
-365.859548389584	-0.523571459585718\\
-364.840441012454	-1.04710304717764\\
-363.821333635324	-1.57055489381855\\
-362.802226258194	-2.09388713662361\\
-361.783118881065	-2.61705992181644\\
-360.764011503935	-3.1400334077638\\
-359.744904126805	-3.66276776801009\\
-358.725796749675	-4.1852231943098\\
-357.706689372545	-4.70735989965957\\
-356.687581995416	-5.2291381213277\\
-355.668474618286	-5.7505181238824\\
-354.649367241156	-6.27146020221799\\
-353.630259864026	-6.7919246845782\\
-352.611152486897	-7.31187193557771\\
-351.592045109767	-7.83126235922021\\
-350.572937732637	-8.35005640191413\\
-349.553830355507	-8.86821455548445\\
-348.534722978378	-9.38569736018172\\
};
\addplot [color=mycolor2, line width=3.0pt, forget plot]
  table[row sep=crcr]{%
348.534722978378	9.38569736018172\\
349.553830355507	8.86821455548445\\
350.572937732637	8.35005640191413\\
351.592045109767	7.83126235922021\\
352.611152486897	7.31187193557771\\
353.630259864026	6.7919246845782\\
354.649367241156	6.27146020221799\\
355.668474618286	5.7505181238824\\
356.687581995416	5.2291381213277\\
357.706689372545	4.70735989965957\\
358.725796749675	4.1852231943098\\
359.744904126805	3.66276776801009\\
360.764011503935	3.1400334077638\\
361.783118881065	2.61705992181644\\
362.802226258194	2.09388713662361\\
363.821333635324	1.57055489381855\\
364.840441012454	1.04710304717764\\
365.859548389584	0.523571459585718\\
366.878655766713	7.34759071599491e-15\\
};
\end{axis}

\definecolor{antiquebrass}{rgb}{0.8, 0.58, 0.46}
\definecolor{darkred}{rgb}{0.7, 0.10, 0.05}
\definecolor{darkgreen}{rgb}{0, 0.5, 0}
\definecolor{darkmagenta}{rgb}{0.7, 0.0, 0.7}

\node[] (bodyframe) at ($(305pt,309pt)$) {};

\draw[->,>=stealth, line width=3pt, color=green,-{Triangle[length=12.0pt,width=7.0pt]}] ($(bodyframe)+(0.0pt,0.0pt)$) -- ($(bodyframe)+(0.0pt,30.0pt)$);
\draw[->,>=stealth, line width=3pt, color=red,-{Triangle[length=12.0pt,width=7.0pt]}] ($(bodyframe)+(0.0pt,0.0pt)$) -- ($(bodyframe)+(30.0pt,0.0pt)$);

\node[overlay, color=black,text width=0.95in] at ($(bodyframe)+(60.0pt,-11.0pt)$) {\scalebox{2.2}{\color{red} $\bm{x}$}};
\node[overlay, color=black,text width=0.95in] at ($(bodyframe)+(15.0pt,25.0pt)$) {\scalebox{2.2}{\color{green} $\bm{y}$}};

\node[overlay, color=black,text width=0.95in] (avebody) at ($(bodyframe)+(-160.0pt,90.0pt)$) {\scalebox{2.5}{\color{brown} $\abProject(s, t)$}};
\draw[->, >=stealth, overlay, line width=0.50mm, solid, color=brown] ($(avebody.south)+(0.0pt,0.0pt)$) -- ($(avebody.south)+(20.0pt,-70.0pt)$);

\node[overlay, color=black,text width=0.95in] (bodycurve) at ($(bodyframe)+(-40.0pt,90.0pt)$) {\scalebox{2.5}{\color{darkgreen} $\gSpace(s, t)$}};
\draw[->, >=stealth, overlay, line width=0.50mm, solid, color=darkgreen] ($(bodycurve.east)+(0.0pt,0.0pt)$) -- ($(bodycurve.east)+(55.0pt,-33.0pt)$);

\node[overlay, color=black,text width=0.95in] (bodygnd) at ($(bodyframe)+(100.0pt,-50.0pt)$) {\scalebox{2.5}{\color{darkmagenta} $\mathcal{S}_{\rm cont}$}};
\draw[->, >=stealth, overlay, line width=0.50mm, solid, color=darkmagenta] ($(bodygnd.north)+(0.0pt,0.0pt)$) -- ($(bodygnd.north)+(78.0pt,42.0pt)$);

\node[overlay, color=black,text width=0.95in] (neg_avebody) at ($(bodyframe)+(-200.0pt,-80.0pt)$) {\scalebox{2.5}{\color{brown} $s = -\frac{L}{2}$}};
\draw[->, >=stealth, overlay, line width=0.50mm, densely dashed, color=brown] ($(neg_avebody.north)+(0.0pt,0.0pt)$) -- ($(neg_avebody.north)+(13.0pt,56.0pt)$);
\node[overlay, color=black,text width=0.95in] (pos_avebody) at ($(bodyframe)+(210.0pt,-80.0pt)$) {\scalebox{2.5}{\color{brown} $s = \frac{L}{2}$}};
\draw[->, >=stealth, overlay, line width=0.50mm, densely dashed, color=brown] ($(pos_avebody.north)+(0.0pt,0.0pt)$) -- ($(pos_avebody.north)+(-20.0pt,56.0pt)$);
\node[overlay, color=black,text width=0.95in] (zero_avebody) at ($(bodyframe)+(0.0pt,-80.0pt)$) {\scalebox{2.5}{\color{brown} $s = 0$}};
\draw[->, >=stealth, overlay, line width=0.50mm, densely dashed, color=brown] ($(zero_avebody.north)+(0.0pt,0.0pt)$) -- ($(zero_avebody.north)+(0.0pt,64.0pt)$);

\end{tikzpicture}%

%% file: dynamics.tex
The shape-centric modeling approach for deriving the locomotive dynamics
of prescribed gaits \cite{ChVe_RAS[2020]} applies to the Turn-in-Place
gait. This section these dynamics to said gait. 

The configuration space, for snake-like robots entails a 
a shape component, $r \in M$, and a group component, $g \in SE(2)$. 
The shape, $r$, is assumed to be fully controlled such that its dynamics
are prescribed by $u$. Modeling emphasis is on the group dynamics of the
robot, as influenced by its shape kinematics and external forces.  
Second-order dynamical equations that focus on the Lie group structure
are 
\begin{equation} \eqlabel{mech_sys_eoms}
  \myvector{ \dot r \\ \dot g \\ \dot p}
  = 
  \myvector{u \\ 
            g \of{\body \Omega - \loc\Pconn(r, u)} \\ 
            \dual\ad_{\of{\body \Omega - \loc\Pconn(r, u)}} p
              + \body{\mathcal{F}}(r,p,u)},
\end{equation}
where $u$ is the shape space control signal. 
The local principle connection $\loc\Pconn(r, u)$ splits the body
velocity, $\xi^b = \inverse{g} \dot g$, into horizontal and vertical
components with $\body \Omega$ denoting the vertical body velocity
component and $p = \lock\li(r) \body \Omega$ the vertical momentum. 
The horizontal body velocity $\loc\Pconn(r, u)$ describes free-space
motion of the body due to shape kinematics.  Vertical components are
driven by external forcing.  The net external wrench,
$\body{\mathcal{F}}$, acts on the body frame and models this external
forcing.  The dual adjoint operation captures Coriolis effects in the
system.

Additional modeling elements must be computed to specialize the generalized dynamics to the Turn-in-Place gait. 
First, the locked inertia tensor associated with the gait is computed as, 
\begin{equation} \eqlabel{gait_locked_inertia}
  \lock{\li}(\gSpace(s)) 
  = \Int[-\frac{L}{2}]{\frac{L}{2}}
        {\partKEM{\ident}{\JJ (s)}{-\gSpace^T(s) \JJ}{\gSpace^T(s) \gSpace(s)}}{s},
\end{equation}
and describes the inertia associated with instantaneous gait shapes,
$\gSpace(s, t)$. Another term denotes dynamic coupling between the robot
shape and group components,
\begin{equation} \eqlabel{gait_princ_conn}
  \KECsym(\gSpace(s), \dot \gSpace(s)) 
    =  \Int[-\frac{L}{2}]{\frac{L}{2}}{\partKEC{R(s)}{-\gSpace^T(s) \JJ R(s)} \dot \gSpace(s)}{s} \!\!\!.
\end{equation} 
$R(s) \in SO(2)$ is the orientation of a local frame $g_{\rm shape}(s)$,
placed at $\gSpace(s)$ with its $x$-axis tangent to the curve, and
\begin{equation} \nonumber
\JJ = \left[ 
 \begin{array}{cc}
 0 & -1 \\
 1 & 0
 \end{array}
 \right ].
\end{equation}

The local connection form $- \loc\Pconn(\gSpace(s), \dot \gSpace(s))$ is computed from these terms and describes free-space motion of the robot body frame due to internal shape changes,
\begin{equation} \eqlabel{conn_form}
\loc\Pconn(\gSpace(s), \dot \gSpace(s)) = -\inverse{\of{\lock{\li}(\gSpace(s))}} \KECsym(\gSpace(s),\dot \gSpace(s)).
\end{equation}
The total body velocity is, $\body \xi = \body \Omega - \loc\Pconn(\gSpace(s), \dot \gSpace(s))$.

\begin{figure*}[t]
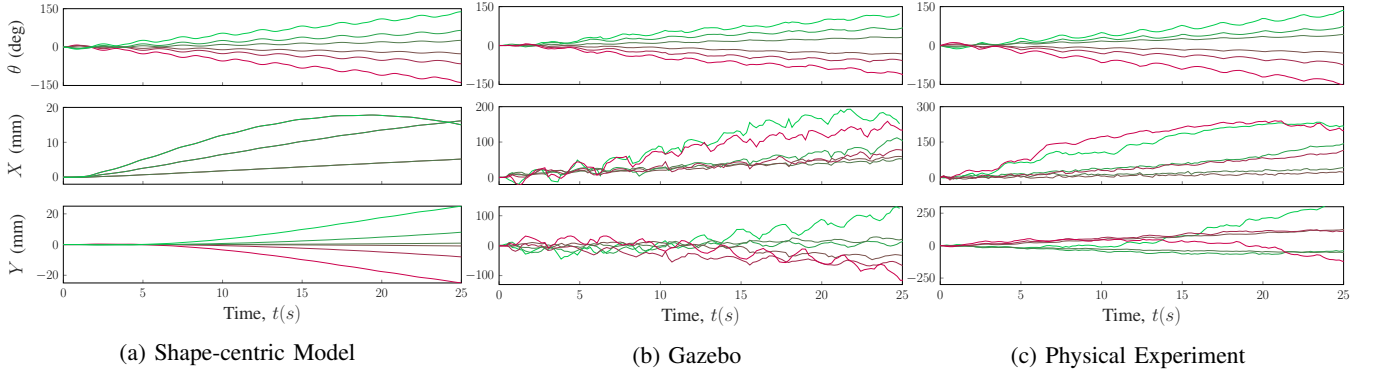

\vspace{0.5mm}
\centering
  \begin{subfigure}{0.71\columnwidth}
	\centering
  \resizebox{1.0\columnwidth}{!}{\input{tip_stepping_matlab_orient.tex}}
	\caption{Shape-centric Model}
	\label{tip_traj_fan_matlab}
	\end{subfigure}
  \hspace{-3.2mm}
  \begin{subfigure}{0.685\columnwidth}
	\centering
  \resizebox{1.0\columnwidth}{!}{\input{tip_stepping_gz_orient.tex}}
	\caption{Gazebo}
	\label{tip_traj_fan_gz}
	\end{subfigure}
  \hspace{-3.2mm}
  \begin{subfigure}{0.685\columnwidth}
	\centering
  \resizebox{1.0\columnwidth}{!}{\input{tip_stepping_exp_orient.tex}}
	\caption{Physical Experiment}
	\label{tip_traj_fan_exp}
	\end{subfigure}
	\caption{Trajectories of the robot are tracked over a series of Turn-in-Place experiments, where the gait's peak amplitude $\bar{A}$ is varied from $-120$ mm (red) to $120$ mm (green). In each experiment, $10$ gait cycles are run. A set of experiments is run in each of: \textbf{(a)} \textit{shape-centric model} numerical integration, \textbf{(b)} \textit{Gazebo simulation} and \textbf{(c)} \textit{physical experiment}. In the latter two, the Turn-in-Place gait is implemented on a $12$-link arculated snake-like robot.} 
	\label{tip_traj_fan_compare}
\end{figure*}
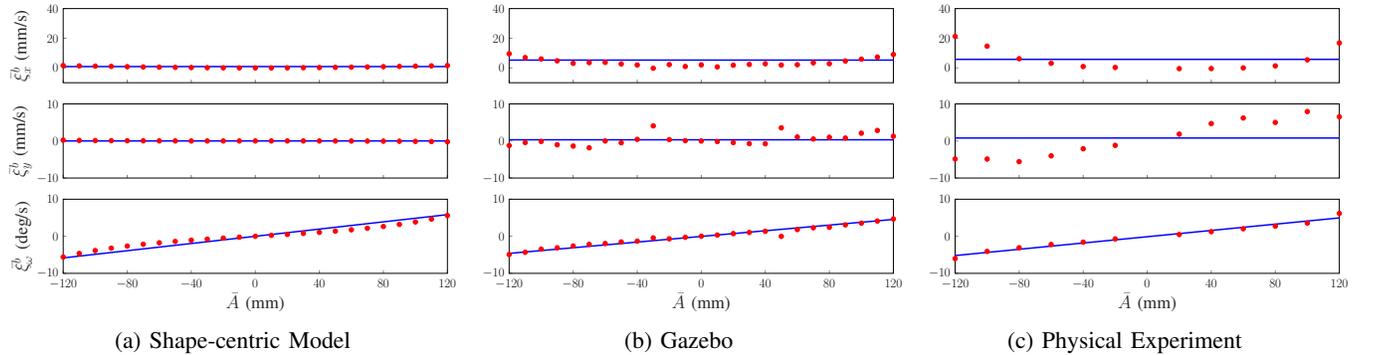
\begin{figure*}[t]
\centering
  \begin{subfigure}{0.69\columnwidth}
	\centering
  \resizebox{1.0\columnwidth}{!}{\input{tip_ctrl2action_map_matlab.tex}}
  	\caption{Shape-centric Model}
	\label{cntrl2act_tip_matlab}
	\end{subfigure}
  \hspace{-3.2mm}
  \begin{subfigure}{0.69\columnwidth}
	\centering
  \resizebox{1.0\columnwidth}{!}{\input{tip_ctrl2action_map_gz.tex}} 
  	\caption{Gazebo}
	\label{cntrl2act_tip_gz}
	\end{subfigure}
  \hspace{-3.2mm}
  \begin{subfigure}{0.69\columnwidth}
	\centering
  \resizebox{1.0\columnwidth}{!}{\input{tip_ctrl2action_map_exp.tex}} 
  	\caption{Physical Experiment}
	\label{cntrl2act_tip_exp}
	\end{subfigure}
	\caption{Time-averaged, steady-behavior body velocities, $\bar{\bm{\xi}}^b$, computed over $20$ gait cycles of the Turn-in-Place gait, for a range of $\bar{A}$. This may alternatively be interpreted as a control-to-action map, denoted $\bm{\Phi}^{\rm TiP}(\bar{A})$. Red markers denote sampled values; blue plots represent linear fits. A set of experiments is run in each of: \textbf{(a)} \textit{shape-centric model} numerical integration, \textbf{(b)} \textit{Gazebo simulation} and \textbf{(c)} \textit{physical experiment}. The latter two, entail a $12$-link arculated snake-like robot.} 
	\label{cntrl2act_tip_compare}
\end{figure*}

\subsection{Body-Environment Forcing}
We apply viscous friction to model robot-environment interactions. Coefficients $\mu_{\rm b}$, $\mu_{\rm f}$ and $\mu_{\rm n}$ denote resistance to posteriorly-, anteriorly- and laterally-directed motion of the body relative to the ground.
At each point of body-ground contact, $s$, frictional forcing is
computed in the local frame $\gshape(s)$ located at $\gSpace(s)$:
\begin{equation} \eqlabel{local_friction_force}
f(s) = \left[ \begin{array}{c} 
-(\mu_{\rm b} H(-v_x(s)) + \mu_{\rm f} H(v_x(s)))v_x(s) \\
-\mu_{\rm n} v_y(s) \end{array} \right].
\end{equation}
$v(s) = \left[ v_x(s) \hspace{3mm} v_y(s) \right]^T$
describes velocity of the body at $\gSpace(s)$ in the local frame $\gshape(s)$.
$H(\phi)$ denotes the Heaviside step function,
\begin{equation}
  H(\phi) = \small \begin{cases}
              0 & \phi \leq 0 \\
              1 & \phi > 0
            \end{cases}.
\end{equation}
Velocity at each point of body-ground contact, $s \in \mathcal{S}_{\rm cont}$, expressed in the local frame at $\gshape(s)$, has contributions from body frame motion and the local shape derivative, 
\begin{equation} \eqlabel{contrib_shape_vel}
\begin{split}
  v(s) = {\rm linear}\of{\inverse\Ad_{\gshape(s)} \body\xi} + R^{-1}_{\rm shape}(s) \dot{r}(s).
\end{split}
\end{equation}

Integration of external forcing acting along the body, after rigid translation to the body frame, 
produces the net wrench acting at the body frame,
\begin{equation} \eqlabel{body_wrench_form}
\begin{split}
  \body{\mathcal{F}} & 
    = \Int[\mathcal{S}_{\rm cont}]{}
        {\of{\inverse\Ad_{\gshape(s)}}^T \cdot f(s)}{s}.
\end{split}
\end{equation}

\subsection{Model Validation}

The terms \eqref{gait_locked_inertia}, \eqref{conn_form} and
\eqref{body_wrench_form} based on the Turn-in-Place gait shape equations
\eqref{gait_shape} and \eqref{tip_contact}, lead to a set of
integro-differential equations for \eqref{mech_sys_eoms}. 
We validate the equations using the Gazebo simulator and physical
experiments, where the Turn-in-Place gait is implemented on a $12$-link
articulated snake-like robot. 
Sweeping over a range of peak amplitude values, $\bar{A}$, and tracking the
resulting body frame trajectories produces the plots in
Figs.~\ref{tip_traj_fan_matlab}, \ref{tip_traj_fan_gz},
and \ref{tip_traj_fan_exp} for the numerical integration of
\eqref{mech_sys_eoms}, Gazebo simulation, and robot implementation.
Trajectories produced across the three experiment classes
illustrate quantitatively consistent orientation trajectories and small
net translation.  Time-averaged body velocity measurements are plotted
in Figure \ref{cntrl2act_tip_compare}; angular velocity trends linearly
and identically across the experiment classes. 

Discrepancies occur in the translational outcomes of Figs.
\ref{tip_traj_fan_compare}a-c (bottom $2$ plots) and Figs.
\ref{cntrl2act_tip_compare}a-c (top $2$ plots). Several circumstances
lead to the differences. 
The articulated snake-like robot used in Gazebo simulation and physical
experiment will have slightly different governing equations relative to
the continuous body model employed in the shape-centric framework
\cite{ChVe_RAS[2020]}.
Additionally, the physical robot is designed to produce anisotropic
friction between its body and ground \cite{ChVe_Robotica[2019]}, which
results in the larger net translational displacements of Figs.~%
\ref{tip_traj_fan_exp} and \ref{cntrl2act_tip_exp}.  Regardless, net
translational displacement in all experiments remains small, less than
half a body length over $10$ gait cycles ($\bar{L} = 800$ mm).

\subsection{Control-to-Action Behavior}
At steady-behavior, the time-averaged body velocity of an undulatory
gait is a useful characterization of its locomotive behavior. 
The time-averaged body velocity was measured for the Turn-in-Place gait
in all $3$ sets of experiments. After reaching a steady-state behavior,
the three body velocity components were time-averaged over several gait
cycles; see results depicted in Fig.~\ref{cntrl2act_tip_compare}.
Time-averaged body velocities were measured (red markers) for a range of
$\bar{A}$ values, with the polynomial fit curve superimposed (blue).

The curve fit provides a closed-form, approximate representation of the
time-averaged body velocity, as a function of $\bar{A}$. Its defines a
control-to-action map denoted as $\bm{\Phi}^{\rm TiP}(\bar{A}):
\mathbb{R} \to \mathbb{R}^3$.  It maps the gait's input parameter,
$\bar{A}$, to the resulting time-averaged body velocity twist $\xi^b$,
and captures the gait's locomotive behavior as a kinematic motion model.
This reduced motion model supports trajectory planning and tracking for
snake-like robots (see \S \secref{plan_track}).

The Gazebo simulator regressed control-to-action map is
\begin{equation} \eqlabel{cntrl2action_TiP}
\bm{\Phi}^{\rm TiP}(\bar{A}) = \mymatrix{5.3 \text{ mm/s} \\ 0 \text{ mm/s}\\ -0.0382 \cdot \bar{A} + 0.0352\text{ deg/s}}.
\end{equation}
It entails a small positive forward bias, trivial lateral velocity, and
linearly varying angular velocity (w.r.t.~$\bar{A}$). 
The range $\bar{A} \in \left[ -120, 120 \right]$ mm reflects saturation
limits based on the shapes achievable by the 12-link articulated
snake-like robot.

\subsection{Shape-centric Modeling}
The Turn-in-Place gait dynamics were derived using a shape-centric
modeling approach \cite{ChVe_RAS[2020]}.  The shape-centric modeling
framework presents a unified approach to
\begin{enumerate}
 \item define continuous-body shape kinematics, and
 \item derive associated locomotive dynamics,
\end{enumerate}
for gaits employed by limbless, elongated-body mobile robots. 
The modeling framework has been used to model a variety of gaits including 
a caterpillar-like rectilinear gait, lateral undulation, and sidewinding 
\cite{ChVe_RAS[2020], ChSeVe_CDC[2016b]}. Here, we demonstrate this
modeling approach additionally describes the locomotive behavior of a
very different gait: a turn-in-place motion primitive.  Characterization
of the gaits' steady-state locomotive behaviors leads to uniformly
structured control-to-action maps. The reduced-order motion
models simplify these gaits' dynamics to differential drive-like
kinematic motion models useful for planning and control.

%% file: tip_ctrl2action_map_matlab.tex
%
%
\begin{tikzpicture}

\begin{axis}[%
width=7.250in,
height=1.400in,
at={(1.213in,3.800in)},
scale only axis,
xmin=-120,
xmax=120,
xtick={-120, -80, -40, 0, 40, 80, 120},
xlabel style={font=\color{white!15!black}},
xticklabels={},
ymin=-10,
ymax=40,
ylabel style={font=\color{white!15!black},yshift=15pt},
ylabel={\Huge $\bar{\xi}^b_{x}$ (mm/s)},
axis background/.style={fill=white},
every tick label/.append style={font=\LARGE},
xticklabel style={yshift=-4pt},
yticklabel style={xshift=-3pt},
xtick pos=left,
ytick pos=left,
]
\addplot [color=blue, forget plot, line width=2.0pt]
  table[row sep=crcr]{%
-150	0.889695310533561\\
-146.969696969697	0.889695310533561\\
-143.939393939394	0.889695310533561\\
-140.909090909091	0.889695310533561\\
-137.878787878788	0.889695310533561\\
-134.848484848485	0.889695310533561\\
-131.818181818182	0.889695310533561\\
-128.787878787879	0.889695310533561\\
-125.757575757576	0.889695310533561\\
-122.727272727273	0.889695310533561\\
-119.69696969697	0.889695310533561\\
-116.666666666667	0.889695310533561\\
-113.636363636364	0.889695310533561\\
-110.606060606061	0.889695310533561\\
-107.575757575758	0.889695310533561\\
-104.545454545455	0.889695310533561\\
-101.515151515152	0.889695310533561\\
-98.4848484848485	0.889695310533561\\
-95.4545454545455	0.889695310533561\\
-92.4242424242424	0.889695310533561\\
-89.3939393939394	0.889695310533561\\
-86.3636363636364	0.889695310533561\\
-83.3333333333333	0.889695310533561\\
-80.3030303030303	0.889695310533561\\
-77.2727272727273	0.889695310533561\\
-74.2424242424242	0.889695310533561\\
-71.2121212121212	0.889695310533561\\
-68.1818181818182	0.889695310533561\\
-65.1515151515152	0.889695310533561\\
-62.1212121212121	0.889695310533561\\
-59.0909090909091	0.889695310533561\\
-56.0606060606061	0.889695310533561\\
-53.030303030303	0.889695310533561\\
-50	0.889695310533561\\
-46.969696969697	0.889695310533561\\
-43.9393939393939	0.889695310533561\\
-40.9090909090909	0.889695310533561\\
-37.8787878787879	0.889695310533561\\
-34.8484848484849	0.889695310533561\\
-31.8181818181818	0.889695310533561\\
-28.7878787878788	0.889695310533561\\
-25.7575757575758	0.889695310533561\\
-22.7272727272727	0.889695310533561\\
-19.6969696969697	0.889695310533561\\
-16.6666666666667	0.889695310533561\\
-13.6363636363636	0.889695310533561\\
-10.6060606060606	0.889695310533561\\
-7.57575757575758	0.889695310533561\\
-4.54545454545455	0.889695310533561\\
-1.51515151515152	0.889695310533561\\
1.51515151515152	0.889695310533561\\
4.54545454545455	0.889695310533561\\
7.57575757575758	0.889695310533561\\
10.6060606060606	0.889695310533561\\
13.6363636363636	0.889695310533561\\
16.6666666666667	0.889695310533561\\
19.6969696969697	0.889695310533561\\
22.7272727272727	0.889695310533561\\
25.7575757575758	0.889695310533561\\
28.7878787878788	0.889695310533561\\
31.8181818181818	0.889695310533561\\
34.8484848484849	0.889695310533561\\
37.8787878787879	0.889695310533561\\
40.9090909090909	0.889695310533561\\
43.9393939393939	0.889695310533561\\
46.969696969697	0.889695310533561\\
50	0.889695310533561\\
53.030303030303	0.889695310533561\\
56.0606060606061	0.889695310533561\\
59.0909090909091	0.889695310533561\\
62.1212121212121	0.889695310533561\\
65.1515151515152	0.889695310533561\\
68.1818181818182	0.889695310533561\\
71.2121212121212	0.889695310533561\\
74.2424242424242	0.889695310533561\\
77.2727272727273	0.889695310533561\\
80.3030303030303	0.889695310533561\\
83.3333333333333	0.889695310533561\\
86.3636363636364	0.889695310533561\\
89.3939393939394	0.889695310533561\\
92.4242424242424	0.889695310533561\\
95.4545454545455	0.889695310533561\\
98.4848484848485	0.889695310533561\\
101.515151515152	0.889695310533561\\
104.545454545455	0.889695310533561\\
107.575757575758	0.889695310533561\\
110.606060606061	0.889695310533561\\
113.636363636364	0.889695310533561\\
116.666666666667	0.889695310533561\\
119.69696969697	0.889695310533561\\
122.727272727273	0.889695310533561\\
125.757575757576	0.889695310533561\\
128.787878787879	0.889695310533561\\
131.818181818182	0.889695310533561\\
134.848484848485	0.889695310533561\\
137.878787878788	0.889695310533561\\
140.909090909091	0.889695310533561\\
143.939393939394	0.889695310533561\\
146.969696969697	0.889695310533561\\
150	0.889695310533561\\
};
\addplot[only marks, mark=*, mark options={}, mark size=3.0000pt, draw=red, fill=red, forget plot] table[row sep=crcr]{%
x	y\\
-150	2.18887812777504\\
-140	1.96966036586464\\
-130	1.7609338504536\\
-120	1.5639148554553\\
-110	1.37149923124033\\
-100	1.18216327000942\\
-90	0.997215945723879\\
-80	0.819046090018085\\
-70	0.650200104407958\\
-60	0.493707488308252\\
-50	0.35303644594328\\
-40	0.231779079090821\\
-30	0.133073453178114\\
-20	0.0600202944302968\\
-10	0.0151487113711818\\
0	0\\
10	0.0151487113711818\\
20	0.0600202944302968\\
30	0.133073453178114\\
40	0.231779079090821\\
50	0.35303644594328\\
60	0.493707488308252\\
70	0.650200104407958\\
80	0.819046090018085\\
90	0.997215945723879\\
100	1.18216327000942\\
110	1.37149923124033\\
120	1.5639148554553\\
130	1.7609338504536\\
140	1.96966036586464\\
150	2.18887812777504\\
};
\end{axis}

\begin{axis}[%
width=7.250in,
height=1.400in,
at={(1.213in,2.000in)},
scale only axis,
xmin=-120,
xmax=120,
xtick={-120, -80, -40, 0, 40, 80, 120},
xlabel style={font=\color{white!15!black}},
xticklabels={},
ymin=-10,
ymax=10,
ytick={-10, 0, 10},
ylabel style={font=\color{white!15!black}},
ylabel={\Huge $\bar{\xi}^b_{y}$ (mm/s)},
axis background/.style={fill=white},
every tick label/.append style={font=\LARGE},
xticklabel style={yshift=-4pt},
yticklabel style={xshift=-3pt},
xtick pos=left,
ytick pos=left,
]
\addplot [color=blue, forget plot, line width=2.0pt]
  table[row sep=crcr]{%
-150	1.22297174697324e-17\\
-146.969696969697	1.22297174697324e-17\\
-143.939393939394	1.22297174697324e-17\\
-140.909090909091	1.22297174697324e-17\\
-137.878787878788	1.22297174697324e-17\\
-134.848484848485	1.22297174697324e-17\\
-131.818181818182	1.22297174697324e-17\\
-128.787878787879	1.22297174697324e-17\\
-125.757575757576	1.22297174697324e-17\\
-122.727272727273	1.22297174697324e-17\\
-119.69696969697	1.22297174697324e-17\\
-116.666666666667	1.22297174697324e-17\\
-113.636363636364	1.22297174697324e-17\\
-110.606060606061	1.22297174697324e-17\\
-107.575757575758	1.22297174697324e-17\\
-104.545454545455	1.22297174697324e-17\\
-101.515151515152	1.22297174697324e-17\\
-98.4848484848485	1.22297174697324e-17\\
-95.4545454545455	1.22297174697324e-17\\
-92.4242424242424	1.22297174697324e-17\\
-89.3939393939394	1.22297174697324e-17\\
-86.3636363636364	1.22297174697324e-17\\
-83.3333333333333	1.22297174697324e-17\\
-80.3030303030303	1.22297174697324e-17\\
-77.2727272727273	1.22297174697324e-17\\
-74.2424242424242	1.22297174697324e-17\\
-71.2121212121212	1.22297174697324e-17\\
-68.1818181818182	1.22297174697324e-17\\
-65.1515151515152	1.22297174697324e-17\\
-62.1212121212121	1.22297174697324e-17\\
-59.0909090909091	1.22297174697324e-17\\
-56.0606060606061	1.22297174697324e-17\\
-53.030303030303	1.22297174697324e-17\\
-50	1.22297174697324e-17\\
-46.969696969697	1.22297174697324e-17\\
-43.9393939393939	1.22297174697324e-17\\
-40.9090909090909	1.22297174697324e-17\\
-37.8787878787879	1.22297174697324e-17\\
-34.8484848484849	1.22297174697324e-17\\
-31.8181818181818	1.22297174697324e-17\\
-28.7878787878788	1.22297174697324e-17\\
-25.7575757575758	1.22297174697324e-17\\
-22.7272727272727	1.22297174697324e-17\\
-19.6969696969697	1.22297174697324e-17\\
-16.6666666666667	1.22297174697324e-17\\
-13.6363636363636	1.22297174697324e-17\\
-10.6060606060606	1.22297174697324e-17\\
-7.57575757575758	1.22297174697324e-17\\
-4.54545454545455	1.22297174697324e-17\\
-1.51515151515152	1.22297174697324e-17\\
1.51515151515152	1.22297174697324e-17\\
4.54545454545455	1.22297174697324e-17\\
7.57575757575758	1.22297174697324e-17\\
10.6060606060606	1.22297174697324e-17\\
13.6363636363636	1.22297174697324e-17\\
16.6666666666667	1.22297174697324e-17\\
19.6969696969697	1.22297174697324e-17\\
22.7272727272727	1.22297174697324e-17\\
25.7575757575758	1.22297174697324e-17\\
28.7878787878788	1.22297174697324e-17\\
31.8181818181818	1.22297174697324e-17\\
34.8484848484849	1.22297174697324e-17\\
37.8787878787879	1.22297174697324e-17\\
40.9090909090909	1.22297174697324e-17\\
43.9393939393939	1.22297174697324e-17\\
46.969696969697	1.22297174697324e-17\\
50	1.22297174697324e-17\\
53.030303030303	1.22297174697324e-17\\
56.0606060606061	1.22297174697324e-17\\
59.0909090909091	1.22297174697324e-17\\
62.1212121212121	1.22297174697324e-17\\
65.1515151515152	1.22297174697324e-17\\
68.1818181818182	1.22297174697324e-17\\
71.2121212121212	1.22297174697324e-17\\
74.2424242424242	1.22297174697324e-17\\
77.2727272727273	1.22297174697324e-17\\
80.3030303030303	1.22297174697324e-17\\
83.3333333333333	1.22297174697324e-17\\
86.3636363636364	1.22297174697324e-17\\
89.3939393939394	1.22297174697324e-17\\
92.4242424242424	1.22297174697324e-17\\
95.4545454545455	1.22297174697324e-17\\
98.4848484848485	1.22297174697324e-17\\
101.515151515152	1.22297174697324e-17\\
104.545454545455	1.22297174697324e-17\\
107.575757575758	1.22297174697324e-17\\
110.606060606061	1.22297174697324e-17\\
113.636363636364	1.22297174697324e-17\\
116.666666666667	1.22297174697324e-17\\
119.69696969697	1.22297174697324e-17\\
122.727272727273	1.22297174697324e-17\\
125.757575757576	1.22297174697324e-17\\
128.787878787879	1.22297174697324e-17\\
131.818181818182	1.22297174697324e-17\\
134.848484848485	1.22297174697324e-17\\
137.878787878788	1.22297174697324e-17\\
140.909090909091	1.22297174697324e-17\\
143.939393939394	1.22297174697324e-17\\
146.969696969697	1.22297174697324e-17\\
150	1.22297174697324e-17\\
};
\addplot[only marks, mark=*, mark options={}, mark size=3.0000pt, draw=red, fill=red, forget plot] table[row sep=crcr]{%
x	y\\
-150	0.669383934579662\\
-140	0.453311666683995\\
-130	0.31360005759163\\
-120	0.220914741575262\\
-110	0.156313433796564\\
-100	0.109918908608146\\
-90	0.0760878425576844\\
-80	0.0513200595994904\\
-70	0.0333218150411926\\
-60	0.0204565308666246\\
-50	0.0116125519018021\\
-40	0.00589798647419753\\
-30	0.00247344006475824\\
-20	0.000726394003855825\\
-10	9.10587888591914e-05\\
0	0\\
10	-9.10587888591914e-05\\
20	-0.000726394003855825\\
30	-0.00247344006475824\\
40	-0.00589798647419753\\
50	-0.0116125519018021\\
60	-0.0204565308666246\\
70	-0.0333218150411926\\
80	-0.0513200595994904\\
90	-0.0760878425576844\\
100	-0.109918908608146\\
110	-0.156313433796564\\
120	-0.220914741575262\\
130	-0.31360005759163\\
140	-0.453311666683995\\
150	-0.669383934579662\\
};
\end{axis}

\begin{axis}[%
width=7.250in,
height=1.400in,
at={(1.213in,0.200in)},
scale only axis,
xmin=-120,
xmax=120,
xtick={-120, -80, -40, 0, 40, 80, 120},
xlabel style={font=\color{white!15!black}},
xlabel={\Huge $\bar{A}$ (mm)},
ymin=-10,
ymax=10,
ytick={-10, 0, 10},
ylabel style={font=\color{white!15!black}},
ylabel={\Huge $\bar{\xi}^b_{\omega}$ (deg/s)},
axis background/.style={fill=white},
every tick label/.append style={font=\LARGE},
xticklabel style={yshift=-4pt},
yticklabel style={xshift=-3pt},
xtick pos=left,
ytick pos=left,
]
\addplot [color=blue, forget plot, line width=2.0pt]
  table[row sep=crcr]{%
-150	-7.29592831363267\\
-146.969696969697	-7.14853582244817\\
-143.939393939394	-7.00114333126368\\
-140.909090909091	-6.85375084007918\\
-137.878787878788	-6.70635834889468\\
-134.848484848485	-6.55896585771018\\
-131.818181818182	-6.41157336652568\\
-128.787878787879	-6.26418087534118\\
-125.757575757576	-6.11678838415668\\
-122.727272727273	-5.96939589297219\\
-119.69696969697	-5.82200340178769\\
-116.666666666667	-5.67461091060319\\
-113.636363636364	-5.52721841941869\\
-110.606060606061	-5.37982592823419\\
-107.575757575758	-5.23243343704969\\
-104.545454545455	-5.0850409458652\\
-101.515151515152	-4.9376484546807\\
-98.4848484848485	-4.7902559634962\\
-95.4545454545455	-4.6428634723117\\
-92.4242424242424	-4.4954709811272\\
-89.3939393939394	-4.3480784899427\\
-86.3636363636364	-4.20068599875821\\
-83.3333333333333	-4.05329350757371\\
-80.3030303030303	-3.90590101638921\\
-77.2727272727273	-3.75850852520471\\
-74.2424242424242	-3.61111603402021\\
-71.2121212121212	-3.46372354283571\\
-68.1818181818182	-3.31633105165121\\
-65.1515151515152	-3.16893856046672\\
-62.1212121212121	-3.02154606928222\\
-59.0909090909091	-2.87415357809772\\
-56.0606060606061	-2.72676108691322\\
-53.030303030303	-2.57936859572872\\
-50	-2.43197610454422\\
-46.969696969697	-2.28458361335973\\
-43.9393939393939	-2.13719112217523\\
-40.9090909090909	-1.98979863099073\\
-37.8787878787879	-1.84240613980623\\
-34.8484848484849	-1.69501364862173\\
-31.8181818181818	-1.54762115743723\\
-28.7878787878788	-1.40022866625273\\
-25.7575757575758	-1.25283617506824\\
-22.7272727272727	-1.10544368388374\\
-19.6969696969697	-0.95805119269924\\
-16.6666666666667	-0.810658701514741\\
-13.6363636363636	-0.663266210330243\\
-10.6060606060606	-0.515873719145744\\
-7.57575757575758	-0.368481227961246\\
-4.54545454545455	-0.221088736776748\\
-1.51515151515152	-0.0736962455922492\\
1.51515151515152	0.0736962455922493\\
4.54545454545455	0.221088736776748\\
7.57575757575758	0.368481227961246\\
10.6060606060606	0.515873719145744\\
13.6363636363636	0.663266210330243\\
16.6666666666667	0.810658701514741\\
19.6969696969697	0.95805119269924\\
22.7272727272727	1.10544368388374\\
25.7575757575758	1.25283617506824\\
28.7878787878788	1.40022866625273\\
31.8181818181818	1.54762115743723\\
34.8484848484849	1.69501364862173\\
37.8787878787879	1.84240613980623\\
40.9090909090909	1.98979863099073\\
43.9393939393939	2.13719112217523\\
46.969696969697	2.28458361335973\\
50	2.43197610454422\\
53.030303030303	2.57936859572872\\
56.0606060606061	2.72676108691322\\
59.0909090909091	2.87415357809772\\
62.1212121212121	3.02154606928222\\
65.1515151515152	3.16893856046672\\
68.1818181818182	3.31633105165121\\
71.2121212121212	3.46372354283571\\
74.2424242424242	3.61111603402021\\
77.2727272727273	3.75850852520471\\
80.3030303030303	3.90590101638921\\
83.3333333333333	4.05329350757371\\
86.3636363636364	4.20068599875821\\
89.3939393939394	4.3480784899427\\
92.4242424242424	4.4954709811272\\
95.4545454545455	4.6428634723117\\
98.4848484848485	4.7902559634962\\
101.515151515152	4.9376484546807\\
104.545454545455	5.0850409458652\\
107.575757575758	5.23243343704969\\
110.606060606061	5.37982592823419\\
113.636363636364	5.52721841941869\\
116.666666666667	5.67461091060319\\
119.69696969697	5.82200340178769\\
122.727272727273	5.96939589297219\\
125.757575757576	6.11678838415668\\
128.787878787879	6.26418087534118\\
131.818181818182	6.41157336652568\\
134.848484848485	6.55896585771018\\
137.878787878788	6.70635834889468\\
140.909090909091	6.85375084007918\\
143.939393939394	7.00114333126368\\
146.969696969697	7.14853582244817\\
150	7.29592831363267\\
};
\addplot[only marks, mark=*, mark options={}, mark size=3.0000pt, draw=red, fill=red, forget plot] table[row sep=crcr]{%
x	y\\
-150	-10.1833683421111\\
-140	-8.2490383523723\\
-130	-6.7531835588896\\
-120	-5.59371357179558\\
-110	-4.64220668621876\\
-100	-3.85793769752333\\
-90	-3.20059778869612\\
-80	-2.64328106726483\\
-70	-2.16405065830002\\
-60	-1.75166886804308\\
-50	-1.38899750254796\\
-40	-1.05910587763514\\
-30	-0.765859036701713\\
-20	-0.499770785782492\\
-10	-0.244886053644276\\
0	0\\
10	0.244886053644276\\
20	0.499770785782492\\
30	0.765859036701713\\
40	1.05910587763514\\
50	1.38899750254796\\
60	1.75166886804308\\
70	2.16405065830002\\
80	2.64328106726483\\
90	3.20059778869612\\
100	3.85793769752333\\
110	4.64220668621876\\
120	5.59371357179558\\
130	6.7531835588896\\
140	8.2490383523723\\
150	10.1833683421111\\
};
\end{axis}
\end{tikzpicture}%

%% file: tip_ctrl2action_map_gz.tex
%
%
\begin{tikzpicture}

\begin{axis}[%
width=7.250in,
height=1.400in,
at={(1.213in,3.800in)},
scale only axis,
xmin=-120,
xmax=120,
xtick={-120, -80, -40, 0, 40, 80, 120},
xticklabels={},
ymin=-10,
ymax=40,
ylabel style={font=\color{white}},
ylabel={\Huge $\bar{\xi}^b_x$ (mm/s)},
axis background/.style={fill=white},
every tick label/.append style={font=\LARGE},
xticklabel style={yshift=-4pt},
yticklabel style={xshift=-3pt},
xtick pos=left,
ytick pos=left,
]
\addplot [color=blue, forget plot, line width=2.0pt]
  table[row sep=crcr]{%
-150	5.26743247643731\\
-146.969696969697	5.26743247643731\\
-143.939393939394	5.26743247643731\\
-140.909090909091	5.26743247643731\\
-137.878787878788	5.26743247643731\\
-134.848484848485	5.26743247643731\\
-131.818181818182	5.26743247643731\\
-128.787878787879	5.26743247643731\\
-125.757575757576	5.26743247643731\\
-122.727272727273	5.26743247643731\\
-119.69696969697	5.26743247643731\\
-116.666666666667	5.26743247643731\\
-113.636363636364	5.26743247643731\\
-110.606060606061	5.26743247643731\\
-107.575757575758	5.26743247643731\\
-104.545454545455	5.26743247643731\\
-101.515151515152	5.26743247643731\\
-98.4848484848485	5.26743247643731\\
-95.4545454545455	5.26743247643731\\
-92.4242424242424	5.26743247643731\\
-89.3939393939394	5.26743247643731\\
-86.3636363636364	5.26743247643731\\
-83.3333333333333	5.26743247643731\\
-80.3030303030303	5.26743247643731\\
-77.2727272727273	5.26743247643731\\
-74.2424242424242	5.26743247643731\\
-71.2121212121212	5.26743247643731\\
-68.1818181818182	5.26743247643731\\
-65.1515151515152	5.26743247643731\\
-62.1212121212121	5.26743247643731\\
-59.0909090909091	5.26743247643731\\
-56.0606060606061	5.26743247643731\\
-53.030303030303	5.26743247643731\\
-50	5.26743247643731\\
-46.969696969697	5.26743247643731\\
-43.9393939393939	5.26743247643731\\
-40.9090909090909	5.26743247643731\\
-37.8787878787879	5.26743247643731\\
-34.8484848484849	5.26743247643731\\
-31.8181818181818	5.26743247643731\\
-28.7878787878788	5.26743247643731\\
-25.7575757575758	5.26743247643731\\
-22.7272727272727	5.26743247643731\\
-19.6969696969697	5.26743247643731\\
-16.6666666666667	5.26743247643731\\
-13.6363636363636	5.26743247643731\\
-10.6060606060606	5.26743247643731\\
-7.57575757575758	5.26743247643731\\
-4.54545454545455	5.26743247643731\\
-1.51515151515152	5.26743247643731\\
1.51515151515152	5.26743247643731\\
4.54545454545455	5.26743247643731\\
7.57575757575758	5.26743247643731\\
10.6060606060606	5.26743247643731\\
13.6363636363636	5.26743247643731\\
16.6666666666667	5.26743247643731\\
19.6969696969697	5.26743247643731\\
22.7272727272727	5.26743247643731\\
25.7575757575758	5.26743247643731\\
28.7878787878788	5.26743247643731\\
31.8181818181818	5.26743247643731\\
34.8484848484849	5.26743247643731\\
37.8787878787879	5.26743247643731\\
40.9090909090909	5.26743247643731\\
43.9393939393939	5.26743247643731\\
46.969696969697	5.26743247643731\\
50	5.26743247643731\\
53.030303030303	5.26743247643731\\
56.0606060606061	5.26743247643731\\
59.0909090909091	5.26743247643731\\
62.1212121212121	5.26743247643731\\
65.1515151515152	5.26743247643731\\
68.1818181818182	5.26743247643731\\
71.2121212121212	5.26743247643731\\
74.2424242424242	5.26743247643731\\
77.2727272727273	5.26743247643731\\
80.3030303030303	5.26743247643731\\
83.3333333333333	5.26743247643731\\
86.3636363636364	5.26743247643731\\
89.3939393939394	5.26743247643731\\
92.4242424242424	5.26743247643731\\
95.4545454545455	5.26743247643731\\
98.4848484848485	5.26743247643731\\
101.515151515152	5.26743247643731\\
104.545454545455	5.26743247643731\\
107.575757575758	5.26743247643731\\
110.606060606061	5.26743247643731\\
113.636363636364	5.26743247643731\\
116.666666666667	5.26743247643731\\
119.69696969697	5.26743247643731\\
122.727272727273	5.26743247643731\\
125.757575757576	5.26743247643731\\
128.787878787879	5.26743247643731\\
131.818181818182	5.26743247643731\\
134.848484848485	5.26743247643731\\
137.878787878788	5.26743247643731\\
140.909090909091	5.26743247643731\\
143.939393939394	5.26743247643731\\
146.969696969697	5.26743247643731\\
150	5.26743247643731\\
};
\addplot[only marks, mark=*, mark options={}, mark size=3.0000pt, draw=red, fill=red, forget plot] table[row sep=crcr]{%
x	y\\
-150	11.621087499172\\
-140	10.4172352319195\\
-130	14.1794367325048\\
-120	9.55511919666916\\
-110	6.93782671286851\\
-100	6.06431776862888\\
-90	4.78281117111568\\
-80	3.0634407166068\\
-70	3.54273352778262\\
-60	3.69330628815568\\
-50	2.65373354216494\\
-40	1.97308909758348\\
-30	-0.242161042187058\\
-20	2.22777603316473\\
-10	0.948590166194329\\
0	2.07382939998176\\
10	0.694217646819007\\
20	1.80079787337415\\
30	2.31119014920383\\
40	2.80440611031981\\
50	1.93857925244189\\
60	2.15770536463207\\
70	3.41277178119745\\
80	2.83239215720634\\
90	4.64697257785144\\
100	5.90533348104184\\
110	7.33786084681507\\
120	9.03512014860243\\
130	12.5843298840279\\
140	10.9468923919705\\
150	11.389665061727\\
};
\end{axis}

\begin{axis}[%
width=7.250in,
height=1.400in,
at={(1.213in,2.000in)},
scale only axis,
xmin=-120,
xmax=120,
xtick={-120, -80, -40, 0, 40, 80, 120},
xticklabels={},
ymin=-10,
ymax=10,
ytick={-10, 0, 10},
ylabel style={font=\color{white}},
ylabel={\Huge $\bar{\xi}^b_y$ (mm/s)},
axis background/.style={fill=white},
every tick label/.append style={font=\LARGE},
xticklabel style={yshift=-4pt},
yticklabel style={xshift=-3pt},
xtick pos=left,
ytick pos=left,
]
\addplot [color=blue, forget plot, line width=2.0pt]
  table[row sep=crcr]{%
-150	0.318820530900887\\
-146.969696969697	0.318820530900887\\
-143.939393939394	0.318820530900887\\
-140.909090909091	0.318820530900887\\
-137.878787878788	0.318820530900887\\
-134.848484848485	0.318820530900887\\
-131.818181818182	0.318820530900887\\
-128.787878787879	0.318820530900887\\
-125.757575757576	0.318820530900887\\
-122.727272727273	0.318820530900887\\
-119.69696969697	0.318820530900887\\
-116.666666666667	0.318820530900887\\
-113.636363636364	0.318820530900887\\
-110.606060606061	0.318820530900887\\
-107.575757575758	0.318820530900887\\
-104.545454545455	0.318820530900887\\
-101.515151515152	0.318820530900887\\
-98.4848484848485	0.318820530900887\\
-95.4545454545455	0.318820530900887\\
-92.4242424242424	0.318820530900887\\
-89.3939393939394	0.318820530900887\\
-86.3636363636364	0.318820530900887\\
-83.3333333333333	0.318820530900887\\
-80.3030303030303	0.318820530900887\\
-77.2727272727273	0.318820530900887\\
-74.2424242424242	0.318820530900887\\
-71.2121212121212	0.318820530900887\\
-68.1818181818182	0.318820530900887\\
-65.1515151515152	0.318820530900887\\
-62.1212121212121	0.318820530900887\\
-59.0909090909091	0.318820530900887\\
-56.0606060606061	0.318820530900887\\
-53.030303030303	0.318820530900887\\
-50	0.318820530900887\\
-46.969696969697	0.318820530900887\\
-43.9393939393939	0.318820530900887\\
-40.9090909090909	0.318820530900887\\
-37.8787878787879	0.318820530900887\\
-34.8484848484849	0.318820530900887\\
-31.8181818181818	0.318820530900887\\
-28.7878787878788	0.318820530900887\\
-25.7575757575758	0.318820530900887\\
-22.7272727272727	0.318820530900887\\
-19.6969696969697	0.318820530900887\\
-16.6666666666667	0.318820530900887\\
-13.6363636363636	0.318820530900887\\
-10.6060606060606	0.318820530900887\\
-7.57575757575758	0.318820530900887\\
-4.54545454545455	0.318820530900887\\
-1.51515151515152	0.318820530900887\\
1.51515151515152	0.318820530900887\\
4.54545454545455	0.318820530900887\\
7.57575757575758	0.318820530900887\\
10.6060606060606	0.318820530900887\\
13.6363636363636	0.318820530900887\\
16.6666666666667	0.318820530900887\\
19.6969696969697	0.318820530900887\\
22.7272727272727	0.318820530900887\\
25.7575757575758	0.318820530900887\\
28.7878787878788	0.318820530900887\\
31.8181818181818	0.318820530900887\\
34.8484848484849	0.318820530900887\\
37.8787878787879	0.318820530900887\\
40.9090909090909	0.318820530900887\\
43.9393939393939	0.318820530900887\\
46.969696969697	0.318820530900887\\
50	0.318820530900887\\
53.030303030303	0.318820530900887\\
56.0606060606061	0.318820530900887\\
59.0909090909091	0.318820530900887\\
62.1212121212121	0.318820530900887\\
65.1515151515152	0.318820530900887\\
68.1818181818182	0.318820530900887\\
71.2121212121212	0.318820530900887\\
74.2424242424242	0.318820530900887\\
77.2727272727273	0.318820530900887\\
80.3030303030303	0.318820530900887\\
83.3333333333333	0.318820530900887\\
86.3636363636364	0.318820530900887\\
89.3939393939394	0.318820530900887\\
92.4242424242424	0.318820530900887\\
95.4545454545455	0.318820530900887\\
98.4848484848485	0.318820530900887\\
101.515151515152	0.318820530900887\\
104.545454545455	0.318820530900887\\
107.575757575758	0.318820530900887\\
110.606060606061	0.318820530900887\\
113.636363636364	0.318820530900887\\
116.666666666667	0.318820530900887\\
119.69696969697	0.318820530900887\\
122.727272727273	0.318820530900887\\
125.757575757576	0.318820530900887\\
128.787878787879	0.318820530900887\\
131.818181818182	0.318820530900887\\
134.848484848485	0.318820530900887\\
137.878787878788	0.318820530900887\\
140.909090909091	0.318820530900887\\
143.939393939394	0.318820530900887\\
146.969696969697	0.318820530900887\\
150	0.318820530900887\\
};
\addplot[only marks, mark=*, mark options={}, mark size=3.0000pt, draw=red, fill=red, forget plot] table[row sep=crcr]{%
x	y\\
-150	-3.27723659765695\\
-140	-1.43142305936857\\
-130	-1.61931135917822\\
-120	-1.22697824575007\\
-110	-0.441250563774035\\
-100	-0.131658897338902\\
-90	-1.02572339983071\\
-80	-1.37874944882095\\
-70	-1.84207246866203\\
-60	-0.00896955014084487\\
-50	-0.495269663890398\\
-40	0.426755838085617\\
-30	4.07500195881849\\
-20	0.361592589396122\\
-10	0.0828496643713898\\
0	-0.0197872148095936\\
10	-0.166269761100509\\
20	-0.44822866314398\\
30	-0.729225463726852\\
40	-0.750713406655434\\
50	3.53868882601979\\
60	1.0834859207725\\
70	0.525928079230544\\
80	1.0004799608475\\
90	0.781692419150584\\
100	2.09462405365523\\
110	2.83943983646031\\
120	1.27146429726923\\
130	2.19361241155095\\
140	1.32973491098185\\
150	3.27095345516542\\
};
\end{axis}

\begin{axis}[%
width=7.250in,
height=1.400in,
at={(1.213in,0.200in)},
scale only axis,
xmin=-120,
xmax=120,
xtick={-120, -80, -40, 0, 40, 80, 120},
xlabel style={font=\color{white!15!black}},
xlabel={\Huge $\bar{A}$ (mm)},
ymin=-10,
ymax=10,
ytick={-10, 0, 10},
ylabel style={font=\color{white}},
ylabel={\Huge $\bar{\xi}^b_\omega$ (deg/s)},
axis background/.style={fill=white},
every tick label/.append style={font=\LARGE},
xticklabel style={yshift=-4pt},
yticklabel style={xshift=-3pt},
xtick pos=left,
ytick pos=left,
]
\addplot [color=blue, forget plot, line width=2.0pt]
  table[row sep=crcr]{%
-150	-5.76452658050554\\
-146.969696969697	-5.64878311648846\\
-143.939393939394	-5.53303965247138\\
-140.909090909091	-5.41729618845431\\
-137.878787878788	-5.30155272443723\\
-134.848484848485	-5.18580926042016\\
-131.818181818182	-5.07006579640308\\
-128.787878787879	-4.95432233238601\\
-125.757575757576	-4.83857886836893\\
-122.727272727273	-4.72283540435186\\
-119.69696969697	-4.60709194033478\\
-116.666666666667	-4.49134847631771\\
-113.636363636364	-4.37560501230063\\
-110.606060606061	-4.25986154828355\\
-107.575757575758	-4.14411808426648\\
-104.545454545455	-4.0283746202494\\
-101.515151515152	-3.91263115623233\\
-98.4848484848485	-3.79688769221525\\
-95.4545454545455	-3.68114422819818\\
-92.4242424242424	-3.5654007641811\\
-89.3939393939394	-3.44965730016403\\
-86.3636363636364	-3.33391383614695\\
-83.3333333333333	-3.21817037212988\\
-80.3030303030303	-3.1024269081128\\
-77.2727272727273	-2.98668344409572\\
-74.2424242424242	-2.87093998007865\\
-71.2121212121212	-2.75519651606157\\
-68.1818181818182	-2.6394530520445\\
-65.1515151515152	-2.52370958802742\\
-62.1212121212121	-2.40796612401035\\
-59.0909090909091	-2.29222265999327\\
-56.0606060606061	-2.1764791959762\\
-53.030303030303	-2.06073573195912\\
-50	-1.94499226794204\\
-46.969696969697	-1.82924880392497\\
-43.9393939393939	-1.71350533990789\\
-40.9090909090909	-1.59776187589082\\
-37.8787878787879	-1.48201841187374\\
-34.8484848484849	-1.36627494785667\\
-31.8181818181818	-1.25053148383959\\
-28.7878787878788	-1.13478801982252\\
-25.7575757575758	-1.01904455580544\\
-22.7272727272727	-0.903301091788364\\
-19.6969696969697	-0.787557627771289\\
-16.6666666666667	-0.671814163754213\\
-13.6363636363636	-0.556070699737138\\
-10.6060606060606	-0.440327235720062\\
-7.57575757575758	-0.324583771702987\\
-4.54545454545455	-0.208840307685911\\
-1.51515151515152	-0.0930968436688356\\
1.51515151515152	0.02264662034824\\
4.54545454545455	0.138390084365315\\
7.57575757575758	0.254133548382391\\
10.6060606060606	0.369877012399467\\
13.6363636363636	0.485620476416542\\
16.6666666666667	0.601363940433617\\
19.6969696969697	0.717107404450693\\
22.7272727272727	0.832850868467769\\
25.7575757575758	0.948594332484844\\
28.7878787878788	1.06433779650192\\
31.8181818181818	1.180081260519\\
34.8484848484849	1.29582472453607\\
37.8787878787879	1.41156818855315\\
40.9090909090909	1.52731165257022\\
43.9393939393939	1.6430551165873\\
46.969696969697	1.75879858060437\\
50	1.87454204462145\\
53.030303030303	1.99028550863852\\
56.0606060606061	2.1060289726556\\
59.0909090909091	2.22177243667267\\
62.1212121212121	2.33751590068975\\
65.1515151515152	2.45325936470683\\
68.1818181818182	2.5690028287239\\
71.2121212121212	2.68474629274098\\
74.2424242424242	2.80048975675805\\
77.2727272727273	2.91623322077513\\
80.3030303030303	3.0319766847922\\
83.3333333333333	3.14772014880928\\
86.3636363636364	3.26346361282635\\
89.3939393939394	3.37920707684343\\
92.4242424242424	3.49495054086051\\
95.4545454545455	3.61069400487758\\
98.4848484848485	3.72643746889466\\
101.515151515152	3.84218093291173\\
104.545454545455	3.95792439692881\\
107.575757575758	4.07366786094588\\
110.606060606061	4.18941132496296\\
113.636363636364	4.30515478898003\\
116.666666666667	4.42089825299711\\
119.69696969697	4.53664171701419\\
122.727272727273	4.65238518103126\\
125.757575757576	4.76812864504834\\
128.787878787879	4.88387210906541\\
131.818181818182	4.99961557308249\\
134.848484848485	5.11535903709956\\
137.878787878788	5.23110250111664\\
140.909090909091	5.34684596513371\\
143.939393939394	5.46258942915079\\
146.969696969697	5.57833289316787\\
150	5.69407635718494\\
};
\addplot[only marks, mark=*, mark options={}, mark size=3.0000pt, draw=red, fill=red, forget plot] table[row sep=crcr]{%
x	y\\
-150	-6.08726820410098\\
-140	-6.00809916312583\\
-130	-5.33776570591697\\
-120	-4.91844043477713\\
-110	-4.32488657449439\\
-100	-3.45893487970239\\
-90	-3.10104850911447\\
-80	-2.60744678123024\\
-70	-2.17870162157346\\
-60	-1.95749526878577\\
-50	-1.52728544226594\\
-40	-1.31071370758424\\
-30	-0.42289061160679\\
-20	-0.684117377751335\\
-10	-0.279885064328275\\
0	0.0100440677849073\\
10	0.349783758620431\\
20	0.719906549273936\\
30	1.03042196539845\\
40	1.35536225384238\\
50	-0.017662109410552\\
60	1.80950404522279\\
70	2.29837168652319\\
80	2.42592087392276\\
90	3.08705786490138\\
100	3.54999802458578\\
110	4.11697332085564\\
120	4.68437551800623\\
130	5.58942904260026\\
140	5.87920579290081\\
150	6.22430822986059\\
};
\end{axis}
\end{tikzpicture}%

%% file: tip_ctrl2action_map_exp.tex
%
%
\begin{tikzpicture}

\begin{axis}[%
width=7.250in,
height=1.400in,
at={(1.213in,3.800in)},
scale only axis,
xmin=-120,
xmax=120,
xtick={-120, -80, -40, 0, 40, 80, 120},
xticklabels={},
ymin=-10,
ymax=40,
ylabel style={font=\color{white}},
ylabel={\Huge $\bar{\xi}^b_x$ (mm/s)},
axis background/.style={fill=white},
every tick label/.append style={font=\LARGE},
xticklabel style={yshift=-4pt},
yticklabel style={xshift=-3pt},
xtick pos=left,
ytick pos=left,
]
\addplot [color=blue, forget plot, line width=2.0pt]
  table[row sep=crcr]{%
-120	5.76037886797324\\
-117.575757575758	5.76037886797324\\
-115.151515151515	5.76037886797324\\
-112.727272727273	5.76037886797324\\
-110.30303030303	5.76037886797324\\
-107.878787878788	5.76037886797324\\
-105.454545454545	5.76037886797324\\
-103.030303030303	5.76037886797324\\
-100.606060606061	5.76037886797324\\
-98.1818181818182	5.76037886797324\\
-95.7575757575758	5.76037886797324\\
-93.3333333333333	5.76037886797324\\
-90.9090909090909	5.76037886797324\\
-88.4848484848485	5.76037886797324\\
-86.0606060606061	5.76037886797324\\
-83.6363636363636	5.76037886797324\\
-81.2121212121212	5.76037886797324\\
-78.7878787878788	5.76037886797324\\
-76.3636363636364	5.76037886797324\\
-73.9393939393939	5.76037886797324\\
-71.5151515151515	5.76037886797324\\
-69.0909090909091	5.76037886797324\\
-66.6666666666667	5.76037886797324\\
-64.2424242424242	5.76037886797324\\
-61.8181818181818	5.76037886797324\\
-59.3939393939394	5.76037886797324\\
-56.969696969697	5.76037886797324\\
-54.5454545454545	5.76037886797324\\
-52.1212121212121	5.76037886797324\\
-49.6969696969697	5.76037886797324\\
-47.2727272727273	5.76037886797324\\
-44.8484848484849	5.76037886797324\\
-42.4242424242424	5.76037886797324\\
-40	5.76037886797324\\
-37.5757575757576	5.76037886797324\\
-35.1515151515151	5.76037886797324\\
-32.7272727272727	5.76037886797324\\
-30.3030303030303	5.76037886797324\\
-27.8787878787879	5.76037886797324\\
-25.4545454545455	5.76037886797324\\
-23.030303030303	5.76037886797324\\
-20.6060606060606	5.76037886797324\\
-18.1818181818182	5.76037886797324\\
-15.7575757575758	5.76037886797324\\
-13.3333333333333	5.76037886797324\\
-10.9090909090909	5.76037886797324\\
-8.48484848484848	5.76037886797324\\
-6.06060606060606	5.76037886797324\\
-3.63636363636364	5.76037886797324\\
-1.21212121212121	5.76037886797324\\
1.21212121212121	5.76037886797324\\
3.63636363636364	5.76037886797324\\
6.06060606060606	5.76037886797324\\
8.48484848484848	5.76037886797324\\
10.9090909090909	5.76037886797324\\
13.3333333333333	5.76037886797324\\
15.7575757575758	5.76037886797324\\
18.1818181818182	5.76037886797324\\
20.6060606060606	5.76037886797324\\
23.030303030303	5.76037886797324\\
25.4545454545455	5.76037886797324\\
27.8787878787879	5.76037886797324\\
30.3030303030303	5.76037886797324\\
32.7272727272727	5.76037886797324\\
35.1515151515151	5.76037886797324\\
37.5757575757576	5.76037886797324\\
40	5.76037886797324\\
42.4242424242424	5.76037886797324\\
44.8484848484849	5.76037886797324\\
47.2727272727273	5.76037886797324\\
49.6969696969697	5.76037886797324\\
52.1212121212121	5.76037886797324\\
54.5454545454545	5.76037886797324\\
56.969696969697	5.76037886797324\\
59.3939393939394	5.76037886797324\\
61.8181818181818	5.76037886797324\\
64.2424242424242	5.76037886797324\\
66.6666666666667	5.76037886797324\\
69.0909090909091	5.76037886797324\\
71.5151515151515	5.76037886797324\\
73.9393939393939	5.76037886797324\\
76.3636363636364	5.76037886797324\\
78.7878787878788	5.76037886797324\\
81.2121212121212	5.76037886797324\\
83.6363636363636	5.76037886797324\\
86.0606060606061	5.76037886797324\\
88.4848484848485	5.76037886797324\\
90.9090909090909	5.76037886797324\\
93.3333333333333	5.76037886797324\\
95.7575757575758	5.76037886797324\\
98.1818181818182	5.76037886797324\\
100.606060606061	5.76037886797324\\
103.030303030303	5.76037886797324\\
105.454545454545	5.76037886797324\\
107.878787878788	5.76037886797324\\
110.30303030303	5.76037886797324\\
112.727272727273	5.76037886797324\\
115.151515151515	5.76037886797324\\
117.575757575758	5.76037886797324\\
120	5.76037886797324\\
};
\addplot[only marks, mark=*, mark options={}, mark size=3.0000pt, draw=red, fill=red, forget plot] table[row sep=crcr]{%
x	y\\
-120	21.2755108219021\\
-100	14.6393513337019\\
-80	6.20222108445047\\
-60	3.12848947028955\\
-40	0.917188696120813\\
-20	0.277376714029505\\
20	-0.490626184085273\\
40	-0.447488203921784\\
60	-0.00963412770488054\\
80	1.35996065063824\\
100	5.4517376329768\\
120	16.8204585272815\\
};
\end{axis}

\begin{axis}[%
width=7.250in,
height=1.400in,
at={(1.213in,2.000in)},
scale only axis,
xmin=-120,
xmax=120,
xtick={-120, -80, -40, 0, 40, 80, 120},
xticklabels={},
ymin=-10,
ymax=10,
ytick={-10, 0, 10},
ylabel style={font=\color{white}},
ylabel={\Huge $\bar{\xi}^b_y$ (mm/s)},
axis background/.style={fill=white},
every tick label/.append style={font=\LARGE},
xticklabel style={yshift=-4pt},
yticklabel style={xshift=-3pt},
xtick pos=left,
ytick pos=left,
]
\addplot [color=blue, forget plot, line width=2.0pt]
  table[row sep=crcr]{%
-120	0.797324019013954\\
-117.575757575758	0.797324019013954\\
-115.151515151515	0.797324019013954\\
-112.727272727273	0.797324019013954\\
-110.30303030303	0.797324019013954\\
-107.878787878788	0.797324019013954\\
-105.454545454545	0.797324019013954\\
-103.030303030303	0.797324019013954\\
-100.606060606061	0.797324019013954\\
-98.1818181818182	0.797324019013954\\
-95.7575757575758	0.797324019013954\\
-93.3333333333333	0.797324019013954\\
-90.9090909090909	0.797324019013954\\
-88.4848484848485	0.797324019013954\\
-86.0606060606061	0.797324019013954\\
-83.6363636363636	0.797324019013954\\
-81.2121212121212	0.797324019013954\\
-78.7878787878788	0.797324019013954\\
-76.3636363636364	0.797324019013954\\
-73.9393939393939	0.797324019013954\\
-71.5151515151515	0.797324019013954\\
-69.0909090909091	0.797324019013954\\
-66.6666666666667	0.797324019013954\\
-64.2424242424242	0.797324019013954\\
-61.8181818181818	0.797324019013954\\
-59.3939393939394	0.797324019013954\\
-56.969696969697	0.797324019013954\\
-54.5454545454545	0.797324019013954\\
-52.1212121212121	0.797324019013954\\
-49.6969696969697	0.797324019013954\\
-47.2727272727273	0.797324019013954\\
-44.8484848484849	0.797324019013954\\
-42.4242424242424	0.797324019013954\\
-40	0.797324019013954\\
-37.5757575757576	0.797324019013954\\
-35.1515151515151	0.797324019013954\\
-32.7272727272727	0.797324019013954\\
-30.3030303030303	0.797324019013954\\
-27.8787878787879	0.797324019013954\\
-25.4545454545455	0.797324019013954\\
-23.030303030303	0.797324019013954\\
-20.6060606060606	0.797324019013954\\
-18.1818181818182	0.797324019013954\\
-15.7575757575758	0.797324019013954\\
-13.3333333333333	0.797324019013954\\
-10.9090909090909	0.797324019013954\\
-8.48484848484848	0.797324019013954\\
-6.06060606060606	0.797324019013954\\
-3.63636363636364	0.797324019013954\\
-1.21212121212121	0.797324019013954\\
1.21212121212121	0.797324019013954\\
3.63636363636364	0.797324019013954\\
6.06060606060606	0.797324019013954\\
8.48484848484848	0.797324019013954\\
10.9090909090909	0.797324019013954\\
13.3333333333333	0.797324019013954\\
15.7575757575758	0.797324019013954\\
18.1818181818182	0.797324019013954\\
20.6060606060606	0.797324019013954\\
23.030303030303	0.797324019013954\\
25.4545454545455	0.797324019013954\\
27.8787878787879	0.797324019013954\\
30.3030303030303	0.797324019013954\\
32.7272727272727	0.797324019013954\\
35.1515151515151	0.797324019013954\\
37.5757575757576	0.797324019013954\\
40	0.797324019013954\\
42.4242424242424	0.797324019013954\\
44.8484848484849	0.797324019013954\\
47.2727272727273	0.797324019013954\\
49.6969696969697	0.797324019013954\\
52.1212121212121	0.797324019013954\\
54.5454545454545	0.797324019013954\\
56.969696969697	0.797324019013954\\
59.3939393939394	0.797324019013954\\
61.8181818181818	0.797324019013954\\
64.2424242424242	0.797324019013954\\
66.6666666666667	0.797324019013954\\
69.0909090909091	0.797324019013954\\
71.5151515151515	0.797324019013954\\
73.9393939393939	0.797324019013954\\
76.3636363636364	0.797324019013954\\
78.7878787878788	0.797324019013954\\
81.2121212121212	0.797324019013954\\
83.6363636363636	0.797324019013954\\
86.0606060606061	0.797324019013954\\
88.4848484848485	0.797324019013954\\
90.9090909090909	0.797324019013954\\
93.3333333333333	0.797324019013954\\
95.7575757575758	0.797324019013954\\
98.1818181818182	0.797324019013954\\
100.606060606061	0.797324019013954\\
103.030303030303	0.797324019013954\\
105.454545454545	0.797324019013954\\
107.878787878788	0.797324019013954\\
110.30303030303	0.797324019013954\\
112.727272727273	0.797324019013954\\
115.151515151515	0.797324019013954\\
117.575757575758	0.797324019013954\\
120	0.797324019013954\\
};
\addplot[only marks, mark=*, mark options={}, mark size=3.0000pt, draw=red, fill=red, forget plot] table[row sep=crcr]{%
x	y\\
-120	-4.86569419063302\\
-100	-4.87975157374232\\
-80	-5.58108042372111\\
-60	-4.01219246436612\\
-40	-2.10969748167624\\
-20	-1.18060456327746\\
20	1.84171869184155\\
40	4.68582356452843\\
60	6.19804964648133\\
80	4.99981049179915\\
100	7.93984138806012\\
120	6.53166514287312\\
};
\end{axis}

\begin{axis}[%
width=7.250in,
height=1.400in,
at={(1.213in,0.200in)},
scale only axis,
xmin=-120,
xmax=120,
xtick={-120, -80, -40, 0, 40, 80, 120},
xlabel style={font=\color{white!15!black}},
xlabel={\Huge $\bar{A}$ (mm)},
ymin=-10,
ymax=10,
ytick={-10, 0, 10},
ylabel style={font=\color{white}},
ylabel={\Huge $\bar{\xi}^b_\omega$ (deg/s)},
axis background/.style={fill=white},
every tick label/.append style={font=\LARGE},
xticklabel style={yshift=-4pt},
yticklabel style={xshift=-3pt},
xtick pos=left,
ytick pos=left,
]
\addplot [color=blue, forget plot, line width=2.0pt]
  table[row sep=crcr]{%
-120	-5.20285999460052\\
-117.575757575758	-5.10018381577648\\
-115.151515151515	-4.99750763695243\\
-112.727272727273	-4.89483145812839\\
-110.30303030303	-4.79215527930435\\
-107.878787878788	-4.6894791004803\\
-105.454545454545	-4.58680292165626\\
-103.030303030303	-4.48412674283222\\
-100.606060606061	-4.38145056400817\\
-98.1818181818182	-4.27877438518413\\
-95.7575757575758	-4.17609820636009\\
-93.3333333333333	-4.07342202753604\\
-90.9090909090909	-3.970745848712\\
-88.4848484848485	-3.86806966988796\\
-86.0606060606061	-3.76539349106391\\
-83.6363636363636	-3.66271731223987\\
-81.2121212121212	-3.56004113341583\\
-78.7878787878788	-3.45736495459179\\
-76.3636363636364	-3.35468877576774\\
-73.9393939393939	-3.2520125969437\\
-71.5151515151515	-3.14933641811966\\
-69.0909090909091	-3.04666023929561\\
-66.6666666666667	-2.94398406047157\\
-64.2424242424242	-2.84130788164753\\
-61.8181818181818	-2.73863170282348\\
-59.3939393939394	-2.63595552399944\\
-56.969696969697	-2.5332793451754\\
-54.5454545454545	-2.43060316635136\\
-52.1212121212121	-2.32792698752731\\
-49.6969696969697	-2.22525080870327\\
-47.2727272727273	-2.12257462987923\\
-44.8484848484849	-2.01989845105518\\
-42.4242424242424	-1.91722227223114\\
-40	-1.8145460934071\\
-37.5757575757576	-1.71186991458305\\
-35.1515151515151	-1.60919373575901\\
-32.7272727272727	-1.50651755693497\\
-30.3030303030303	-1.40384137811092\\
-27.8787878787879	-1.30116519928688\\
-25.4545454545455	-1.19848902046284\\
-23.030303030303	-1.09581284163879\\
-20.6060606060606	-0.993136662814752\\
-18.1818181818182	-0.890460483990709\\
-15.7575757575758	-0.787784305166666\\
-13.3333333333333	-0.685108126342623\\
-10.9090909090909	-0.58243194751858\\
-8.48484848484848	-0.479755768694537\\
-6.06060606060606	-0.377079589870494\\
-3.63636363636364	-0.274403411046451\\
-1.21212121212121	-0.171727232222407\\
1.21212121212121	-0.0690510533983644\\
3.63636363636364	0.0336251254256787\\
6.06060606060606	0.136301304249722\\
8.48484848484848	0.238977483073765\\
10.9090909090909	0.341653661897808\\
13.3333333333333	0.444329840721851\\
15.7575757575758	0.547006019545894\\
18.1818181818182	0.649682198369937\\
20.6060606060606	0.75235837719398\\
23.030303030303	0.855034556018023\\
25.4545454545455	0.957710734842066\\
27.8787878787879	1.06038691366611\\
30.3030303030303	1.16306309249015\\
32.7272727272727	1.2657392713142\\
35.1515151515151	1.36841545013824\\
37.5757575757576	1.47109162896228\\
40	1.57376780778632\\
42.4242424242424	1.67644398661037\\
44.8484848484849	1.77912016543441\\
47.2727272727273	1.88179634425845\\
49.6969696969697	1.9844725230825\\
52.1212121212121	2.08714870190654\\
54.5454545454545	2.18982488073058\\
56.969696969697	2.29250105955463\\
59.3939393939394	2.39517723837867\\
61.8181818181818	2.49785341720271\\
64.2424242424242	2.60052959602676\\
66.6666666666667	2.7032057748508\\
69.0909090909091	2.80588195367484\\
71.5151515151515	2.90855813249888\\
73.9393939393939	3.01123431132293\\
76.3636363636364	3.11391049014697\\
78.7878787878788	3.21658666897101\\
81.2121212121212	3.31926284779506\\
83.6363636363636	3.4219390266191\\
86.0606060606061	3.52461520544314\\
88.4848484848485	3.62729138426719\\
90.9090909090909	3.72996756309123\\
93.3333333333333	3.83264374191527\\
95.7575757575758	3.93531992073932\\
98.1818181818182	4.03799609956336\\
100.606060606061	4.1406722783874\\
103.030303030303	4.24334845721144\\
105.454545454545	4.34602463603549\\
107.878787878788	4.44870081485953\\
110.30303030303	4.55137699368357\\
112.727272727273	4.65405317250762\\
115.151515151515	4.75672935133166\\
117.575757575758	4.8594055301557\\
120	4.96208170897975\\
};
\addplot[only marks, mark=*, mark options={}, mark size=3.0000pt, draw=red, fill=red, forget plot] table[row sep=crcr]{%
x	y\\
-120	-6.00724494350607\\
-100	-4.08016984100807\\
-80	-3.08598150458782\\
-60	-2.21254753191912\\
-40	-1.58500614184543\\
-20	-0.709215297195201\\
20	0.47865530683246\\
40	1.24772101888134\\
60	2.03575070121263\\
80	2.74586645114967\\
100	3.56666021916837\\
120	6.1608418490926\\
};
\end{axis}
\end{tikzpicture}%

%% file: plan_track.tex
Multi-gait trajectory synthesis is accomplished using a three-pass
approach to output a discrete-time reference trajectory with
feed-forward controls.  Each pass after the first incrementally refines
the current trajectory.  The final product is a controlled,
switched-gait trajectory designed to move the robot from its starting
pose to a prescribed goal position.  Synthesis is specialized to the two
gaits, Rectilinear and Turn-in-Place, $\mathcal{MP} = \{ {\rm RL}, {\rm
TiP} \}$.  The Rectilinear gait employs a caterpillar-like,
anteriorly-propagating body wave to produce (predominantly)
translational motion. Limited turning ability is accomplished by
modulating the average (reference) body curvature $\kappa$
\cite{ChVe_RAS[2020]}. Its control-to-action map in the Gazebo
simulation environment is, 
\begin{equation} \eqlabel{control_to_action_map_RL}
  \bm{\Phi}^{\rm RL}(\kappa) = \mymatrix{36 \text{ mm/s}, 0 \text{ mm/s}, \text{-}1696.7 \hspace{-1pt} \cdot \hspace{-1pt} \kappa \text{ deg/s}}^T, 
\end{equation}
for $\kappa \in \left[ -1.5e\text{-}3, 1.5e\text{-}3 \right]$ mm$^{-1}$. Other gait parameters associated with the Rectilinear gait remain fixed: frequency ($f^{\rm RL} = -0.4$ Hz), body wave amplitude ($A^{\rm RL} = 60$ mm), wavelength ($\lambda^{\rm RL} = 390$ mm).
The Turn-in-Place control-to-action map is \eqref{cntrl2action_TiP}. 
The map captures locomotive behavior as a kinematic motion model,
parametrized with respect to peak amplitude $\bar{A}$.

Trajectory synthesis starts with a collision-free input seed path
$\bm{G}^{\rm seed}$ consisting of a sequence of coordinates connecting
the robot's start pose and goal position, typically generated using
one's traditional path planning approach \cite{Latombe_RoboMP[1991]}
of choice.  The path generated in this first pass is absent of any
control or timing information and is not constrained by the robot's
locomotive dynamics. 

\subsection{Pass 2: Unicycle-based Trajectory Synthesis}
\seclabel{MPC_traj_synth}
Pass 2 parametrizes the path $\bm{G}^{\rm seed}$ with respect to time.
Presuming a fixed arc length speed consistent with the Rectilinear gait,
the path is re-sampled at a fixed number of Rectilinear gait periods,
$T^{\rm intvl} = \bar{N}^{\rm RL}/{f^{\rm RL}}$, where $\bar{N}^{\rm RL}
= 2$. 
This discrete-time path is the reference path for a model predictive
control (MPC) output tracker based on a unicycle motion model whose
inputs are linear and angular velocities, $v^{\rm uni}$ and $\omega^{\rm uni}$.
Additional constraints limit the unicycle model to motion that is
collectively achievable by the gaits in $\mathcal{MP}$,
\begin{equation}
v^{\rm uni} \in \left[ \Phi^{\rm RL}_x(\kappa) - \epsilon, \Phi^{\rm
RL}_x(\kappa) + \epsilon \right]\ \land\  
\omega^{\rm uni} \in \left[ \omega^{\rm TiP}_{\rm min}, \ \omega^{\rm TiP}_{\rm max} \right]
\end{equation}
where
\begin{equation}
\omega^{\rm TiP}_{\rm min} = \min_{\bar{A}}\Phi^{\rm TiP}_\omega \left(
\bar{A} \right)\ \text{and}\ 
\omega^{\rm TiP}_{\rm max} = \max_{\bar{A}} \Phi^{\rm TiP}_\omega \left( \bar{A} \right).
\end{equation}
The small constant $\epsilon$ adds slack to assist with convergence to a
solution. In summary, linear velocity is constrained to that achievable
by the Rectilinear gait, while the Turn-in-Place gait defines allowable
angular velocities.

The output is 
$\left( \bm{G}^{\rm uni}, \bm{t}^{\rm uni}, \bm{U}^{\rm uni} \right)$,
a collection of sets each of which has cardinality $N^{\rm uni}$.
$\bm{G}^{\rm uni} = \left\{ g^{\rm uni}_i \in SE(2) \right\}^{N^{\rm
uni}}_{i = 1}$ is a sequence of pose coordinates. 
A corresponding time schedule $\bm{t}^{\rm uni}$ consists of timestamps
at intervals of $T^{\rm intvl}$ seconds. 
Lastly, $\bm{U}^{\rm uni} = \left\{ \left( v^{\rm uni}_i, w^{\rm uni}_i
\right) \right\}^{N^{\rm uni}}_{i = 1}$ are the open-loop control
inputs for achieving the pose sequence $\bm{G}^{\rm uni}$. 

\subsection{Pass 3: Multi-gait Refinement}
Pass 3 further revises the sets based on which gaits in $\mathcal{MP}$
will realize the desired displacements.
Stepping through each trajectory point $i = 1 \ldots
N^{\rm uni}$, the ability of the Rectilinear gait to apply the
associated open-loop control is tested:
\begin{equation} \eqlabel{w_thresh_RL_TiP}
  {\rm isRLTraversable}(i) = \left( \omega^{\rm uni}_i \leq 0.85 \cdot \omega^{\rm RL}_{\rm max} \right) \\
\end{equation}
for the maximum angular velocity of the Rectilinear gait,
\begin{equation}
  \omega^{\rm RL}_{\rm max} = \max_{\kappa} \Phi^{\rm RL}_\omega \left( \kappa \right).
\end{equation}
Trajectory points satisfying \eqref{w_thresh_RL_TiP} entail feasible angular
velocities $\omega^{\rm uni}_i$ for the Rectilinear gait using
an average (reference) body curvature
$\kappa_i = \left(\Phi^{\rm RL}_\omega \right)^{-1} \left(\omega^{\rm uni}_i\right)$. 

Trajectory segments where \eqref{w_thresh_RL_TiP} is false are presumed
infeasible for the Rectilinear gait. The robot will employ a
Turn-in-Place, Rectilinear, Turn-in-Place gait sequence to negotiate
them. A trajectory segment \textit{not} satisfying \eqref{w_thresh_RL_TiP}, beginning at point $j$ and ending at point $k > j$, is circumvented by connecting a straight line between the two points; we denote this line's spatial orientation as $\theta_{j \rightarrow k}$. At point $j$, a series of $6$ Turn-in-Place gait cycles (trajectory points) is inserted to close the error, $\theta_{j \rightarrow k} - \theta^{\rm uni}_{j}$, with an angular velocity $\omega^{\rm MG} = \frac{(\theta_{j \rightarrow k} - \theta^{\rm uni}_{j}) \cdot f^{\rm TiP}}{6}$. Following this, we insert a series of Rectilinear gait cycles to traverse the distance from $g^{\rm uni}_{j}$ to $g^{\rm uni}_k$, with the Rectilinear turning parameter fixed to $\kappa = 0$; the number of Rectilinear gait cycles inserted is dictated by the fixed speed at which the gait travels, $\Phi^{\rm RL}_x(\kappa)$. 
At $g^{\rm uni}_k$, a final sequence of $6$ Turn-in-Place gait cycles is inserted to close the remaining orientation error, $\theta^{\rm uni}_{k} - \theta_{j \rightarrow k}$, with angular velocity $\omega^{\rm MG} = \frac{(\theta^{\rm uni}_{k} - \theta_{j \rightarrow k}) \cdot f^{\rm TiP}}{6}$. Turn-in-Place input parameters needed to achieve a desired angular velocity are $\bar{A} = \left( \Phi^{\rm TiP}_\omega \right)^{-1} \left(\omega\right)$.

The refined multi-gait trajectory is described by the collection, $\left( \bm{G}^{\rm MG}, \bm{t}^{\rm MG}, \bm{M\!P}^{\rm MG}, \bm{U}^{\rm MG} \right)$. Each set in the collection is of cardinality $N^{\rm MG}$, the total number of points in the refined multi-gait trajectory. $\bm{G}^{\rm MG}$ and $\bm{t}^{\rm MG}$ denote planned pose coordinates and time, respectively. $\bm{M\!P}^{\rm MG} = \left\{ mp^{\rm MG}_i \in \mathcal{MP} \right\}^{N^{\rm MG}}_{i=1}$ expresses the gait sequence designed to traverse $\bm{G}^{\rm MG}$, while $\bm{U}^{\rm MG} = \left\{ \beta^{\rm MG}_i \in \mathbb{R} \right\}^{N^{\rm MG}}_{i=1}$ contains the gait-specific input parameter to be commanded at each trajectory point. $\beta^{\rm MG}_i$ describes either Rectilinear curvature $\kappa^{\rm MG}_i$ or Turn-in-Place peak amplitude $\bar{A}^{\rm MG}_i$, when $mp^{\rm MG}_i = \text{RL or TiP}$, respectively.

\subsection{Trajectory Tracking}
Tracking of the multi-gait trajectory is accomplished using simple
proportional feedback control. Feedback control is computed at scheduled
time intervals, specified in $\bm{t}^{\rm MG}$. We denote the waypoint
being tracked by its index $i = 1 \ldots N^{\rm MG}$. The pose error
associated with any waypoint $g^{\rm MG}_i \in \bm{G}^{\rm MG}$ and
robot pose $g$ is $g_{\rm err} = g^{-1} \cdot g^{\rm MG}_i$. It is the
the waypoint frame $g^{\rm MG}_i$ relative to the actual robot
frame. Individual translational and rotational errors, 
$\left( x_{\rm err}, y_{\rm err}, \theta_{\rm err} \right)$, are extracted.

Feedback-correction is dependent upon the planned gait $mp_{i}$, at a each waypoint $i$,
\begin{equation}
\omega_{\rm fb} = 
\begin{cases}
k^{\rm RL}_{\theta} \cdot \theta_{\rm err} + k^{\rm RL}_{y} \cdot y_{\rm err}, & mp_{i} = \text{RL} \\
k^{\rm TiP}_{\theta} \cdot \theta_{\rm err}, & mp_{i} = \text{TiP} 
\end{cases}.
\end{equation}
The open-loop gait parameter, $\kappa^{\rm MG}_i$ or $\bar{A}^{\rm
MG}_i$, is incorporated as a feed-forward value in the applied control,
\begin{align}
\kappa_{\rm fb} = \left( \Phi^{RL}_{\omega} \right)^{-1} \left( \Phi^{RL}_{\omega}\left( \kappa^{\rm MG}_{i} \right) + \omega_{\rm fb} \right), & \ mp_{i} = \text{RL} \\
\bar{A}_{\rm fb} =  \left( \Phi^{TiP}_{\omega} \right)^{-1} \left( \Phi^{TiP}_{\omega}\left( \bar{A}^{\rm MG}_i \right) + \omega_{\rm fb} \right), & \ mp_{i} = \text{TiP} 
\end{align}

As the robot proceeds sequentially through the planned waypoints, in $\bm{G}^{\rm MG}$, the forward error is assessed. If it exceeds a threshold, $x_{\rm err} > \bar{d}_{\rm err}$, the relevant target waypoint $i$ is \textit{not} incremented; the robot continues tracking the same waypoint until forward error falls below $\bar{d}_{\rm err}$.

%% file: results.tex
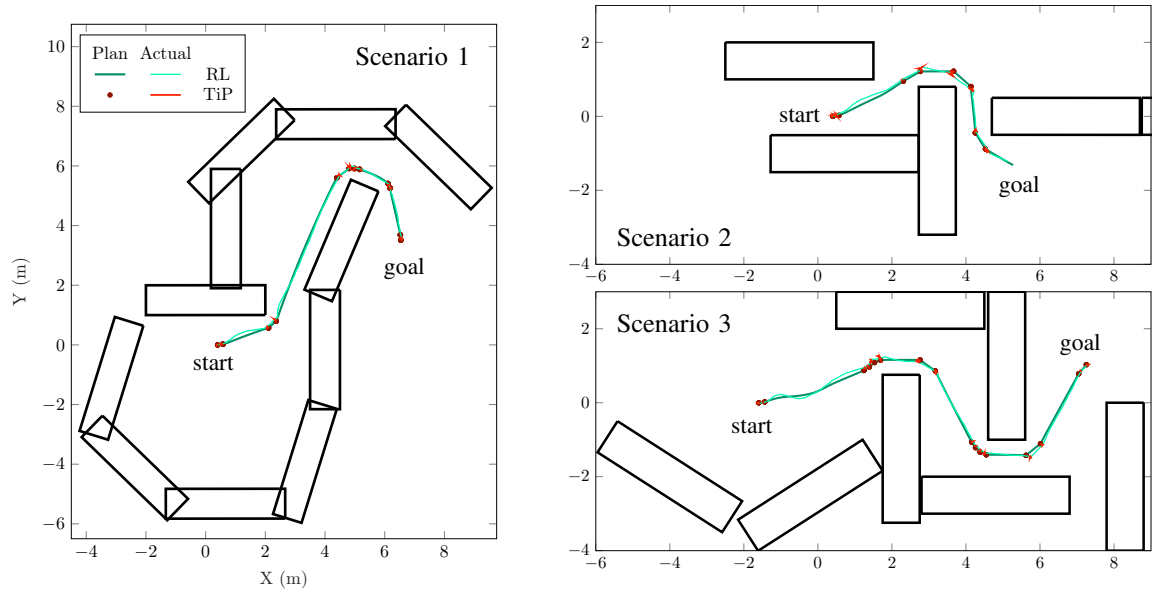
\begin{figure*}[t]
  \vspace{1.0mm}
	\centering
  \begin{tikzpicture}[inner sep=0pt, outer sep=0pt]
  	\def\firstScenWidth{0.75\columnwidth}
  	\def\secScenWidth{0.9\columnwidth}
  	\def\thirdScenWidth{0.9\columnwidth}
  	\def\secScenHorizPos{0.85\columnwidth}
  	\def\labScenHorizPos{0.925\columnwidth}
  
    	\node[anchor=north west,outer sep=0pt,minimum height=3in,minimum width=\firstScenWidth,draw=none] (scen1_map) at (0in,3in) 
    	{\includegraphics[width=\firstScenWidth]{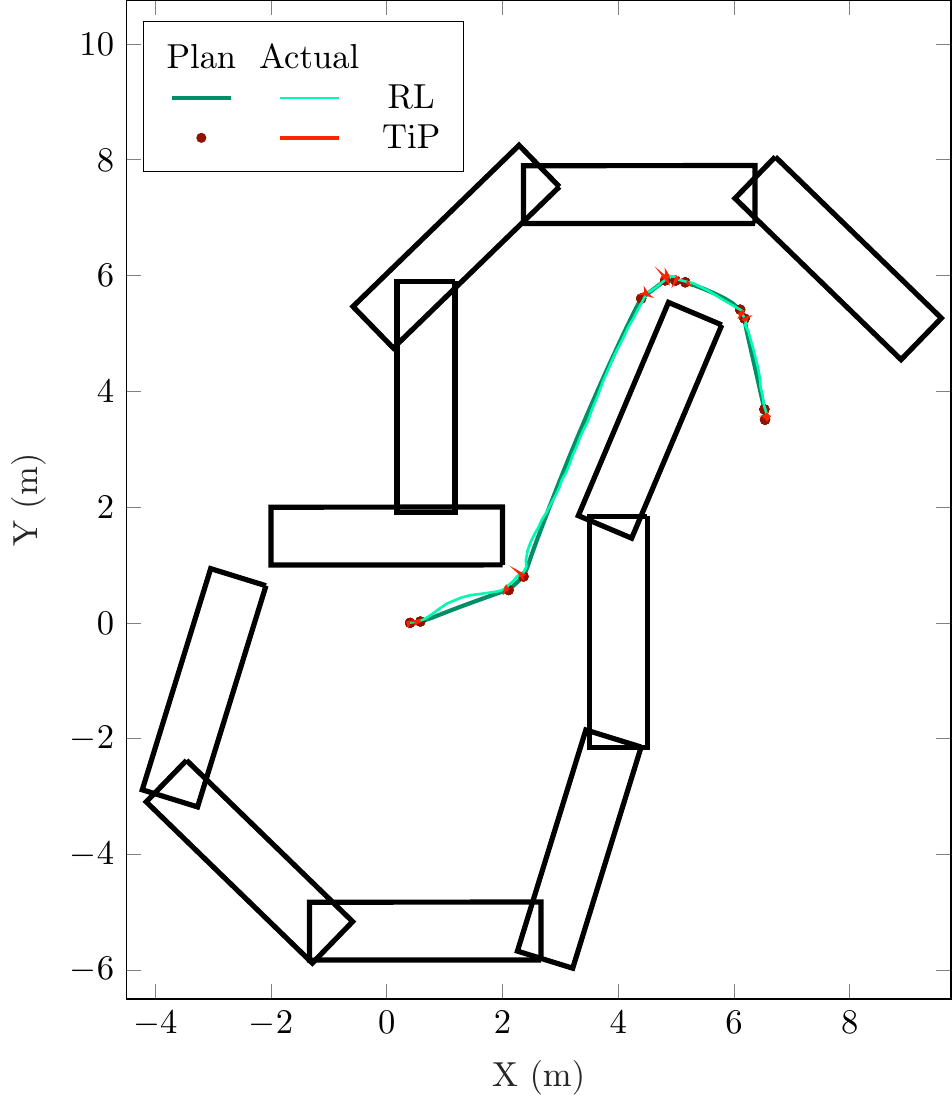}};
    \node[anchor=north west,outer sep=0pt,minimum height=1.5in,minimum width=\secScenWidth,draw=none] (scen2_map) at ($(\secScenHorizPos,3in)+(0in,0.1in)$)
    	{\includegraphics[width=\secScenWidth]{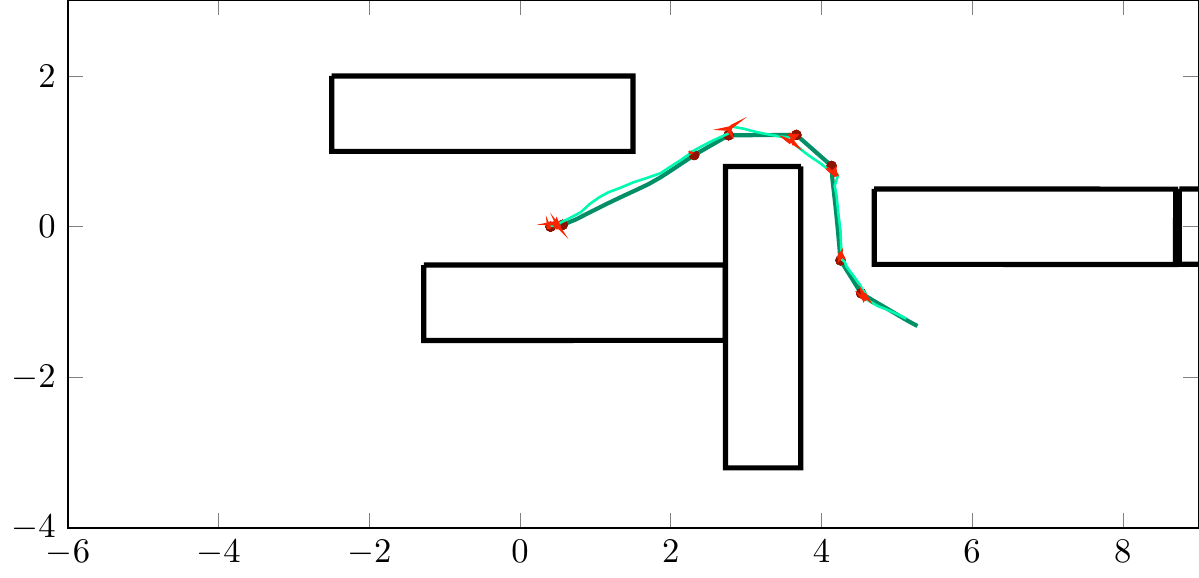}};
    \node[anchor=north west,outer sep=0pt,minimum height=1.5in,minimum width=\thirdScenWidth,draw=none] (scen3_map) at ($(\secScenHorizPos,1.5in)+(0in,0.1in)$)
    	{\includegraphics[width=\thirdScenWidth]{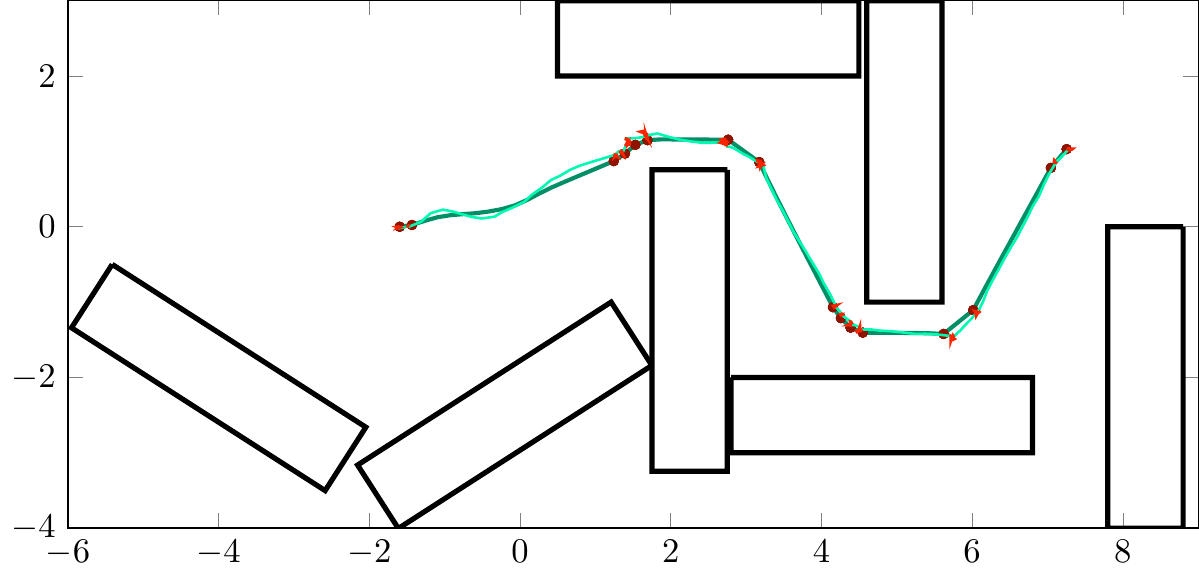}};

    \node[anchor=north west,outer sep=2pt] at (1.8in,2.9in) {Scenario 1};
    \node[anchor=north west,outer sep=2pt] at (\labScenHorizPos,1.95in) {Scenario 2};
    \node[anchor=north west,outer sep=2pt] at (\labScenHorizPos,1.5in) {Scenario 3};

    \node[anchor=north] at ($(scen1_map)+(-0.20in,-0.25in)$) {\small start};
    \node[anchor=north] at ($(scen2_map)+(-0.30in,0.20in)$) {\small start};
    \node[anchor=north] at ($(scen3_map)+(-0.55in,0.07in)$) {\small start};
    \node[anchor=north] at ($(scen1_map)+(0.80in,0.27in)$) {\small goal};
    \node[anchor=north] at ($(scen2_map)+(0.85in,-0.15in)$) {\small goal};
    \node[anchor=north] at ($(scen3_map)+(1.17in,0.52in)$) {\small goal};
  \end{tikzpicture}
  \caption{Depiction of environments for the three locomotion scenarios,
    with parametric plots of the multi-gait planned and achieved
    (based on tracking) trajectories. The snake-like robot
    successfully maneuvers to the goal in all
    cases.\label{fig:mult-gait-scen}} 
  \vspace*{-1em}
\end{figure*}

This section pairs the Turn-in-Place gait with a translation-dominant
Rectilinear gait and show that a $12$-link snake-like robot
can plan and track trajectories that neither gait alone is capable of
following.  
Three locomotion scenarios, depicted in Fig.~\ref{fig:mult-gait-scen},
are successfully completed using these two gaits within the multi-gait
trajectory planning and tracking approach presented in
\S\secref{plan_track}.  Rectilinear motion is used for path segments
that require little steering.  The multi-gait planner switches to
Turn-in-Place motion at ``sharp'' corners and is used at the start of
each trajectory to align the robot body frame with the direction of
travel.

Taken together the scenarios require sufficiently diverse motion
profiles that they can be deemed reflective of most navigation scenarios
with comparable free space vs collision space characteristics.
The narrowest passageways have widths equivalent to the robot body
length ($\bar{L} = 800$ mm).  
Tracking errors arise due to initial transients when momentum
accumulates, transitions when gait parameter values change, and
gait switching. These effects are un-modeled and are not considered
as part of the multi-gait trajectory synthesis process.  Discrete-time,
trajectory tracking control corrects for these errors.  

\begin{figure*}[t]
  \vspace{0.5mm}
  \begin{center}
  \scalebox{0.95}{
  \begin{tikzpicture}[inner sep=0pt, outer sep=0pt]
    	\node[draw=none,rectangle,minimum width=\columnwidth,minimum height=0.75in,anchor=south west] (B) at (0in,0in)
    {\includegraphics[width=\columnwidth]{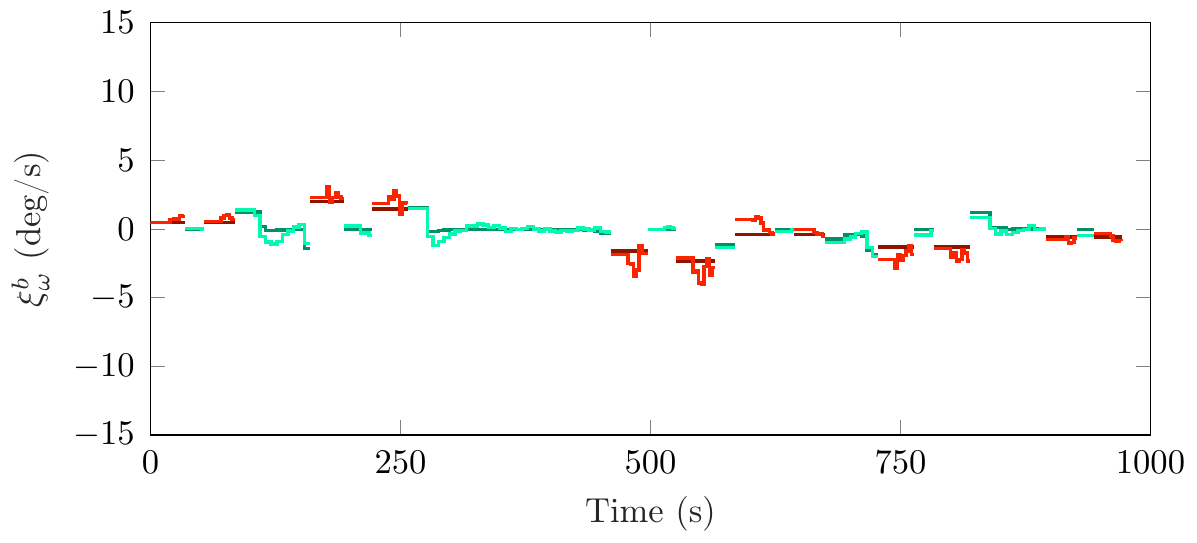}};
    	\node[draw=none,rectangle,minimum width=\columnwidth,minimum height=0.75in,anchor=south west] (A) at ($(B.north west)+(-1.5mm,0in)$)
    {\includegraphics[width=0.98\columnwidth]{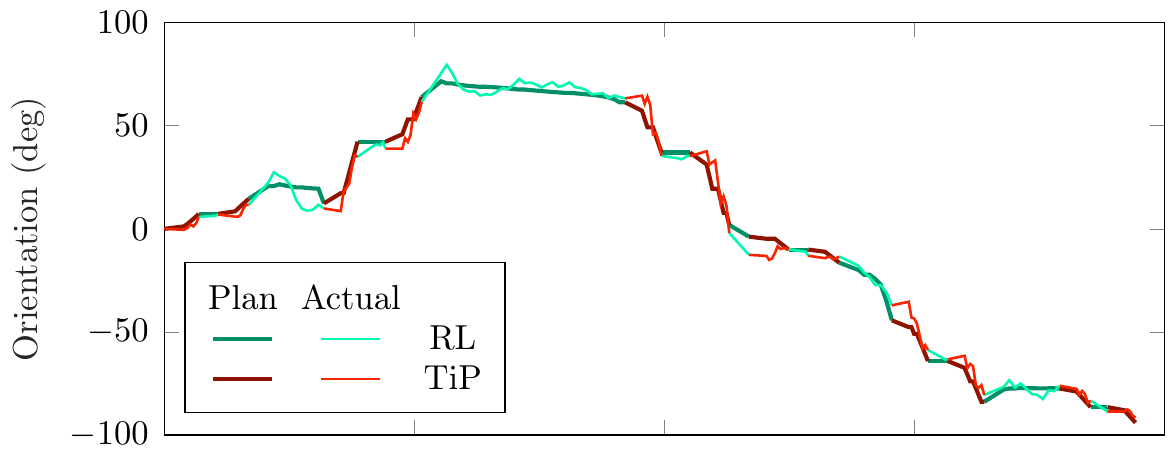}};
    \node[draw=none,rectangle,minimum width=\columnwidth,minimum
    height=0.75in,anchor=south west,xshift=5pt] (D) at (B.south east)
    {\includegraphics[width=\columnwidth]{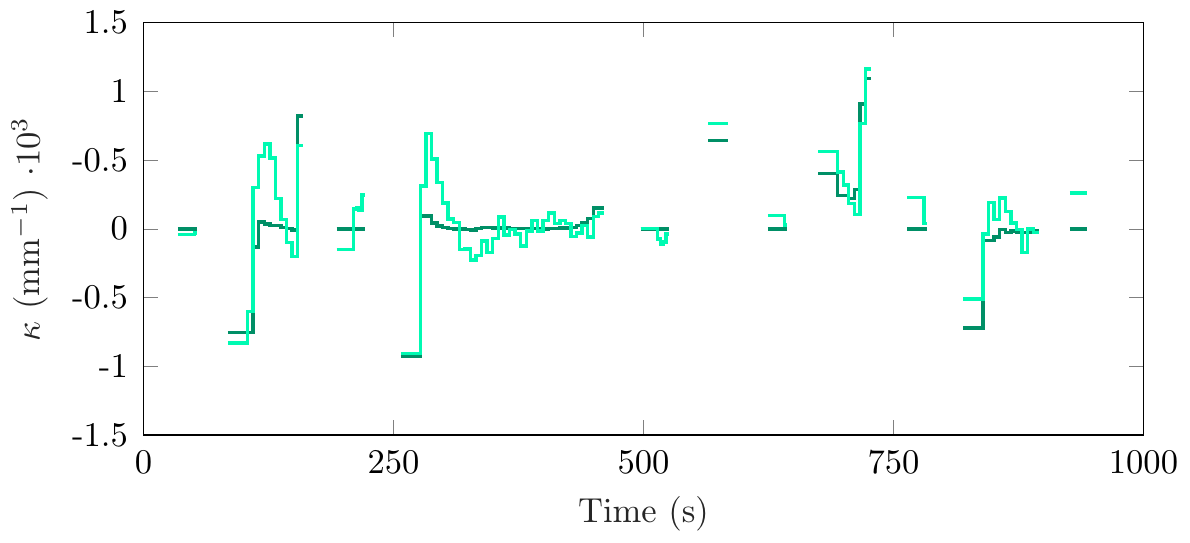}};
    \node[draw=none,rectangle,minimum width=\columnwidth,minimum
    height=0.75in,anchor=south west,xshift=5pt] (C) at ($(A.south east)+(-0.5mm,0in)$)
    {\includegraphics[width=0.99\columnwidth]{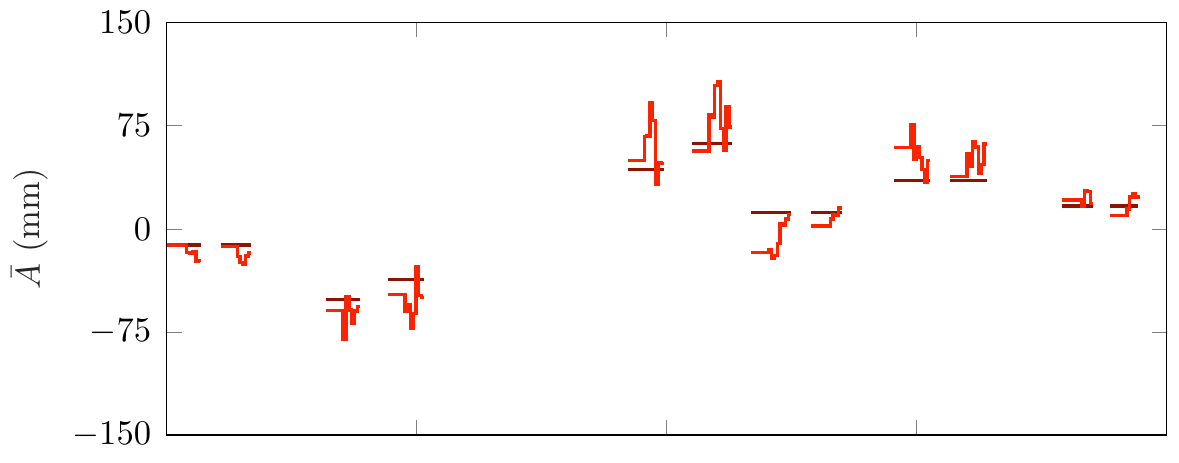}};
  \end{tikzpicture}}
  \vspace{-2.0mm}
  \subcaption{Scenario 1 signals}
  \vspace{3.0mm}
  \centering
  \scalebox{0.95}{
  \begin{tikzpicture}[inner sep=0pt, outer sep=0pt]
    \node[draw=none,rectangle,minimum width=\columnwidth,minimum height=0.75in,anchor=south west] (B) at (0in,0in)
    {\includegraphics[width=0.99\columnwidth]{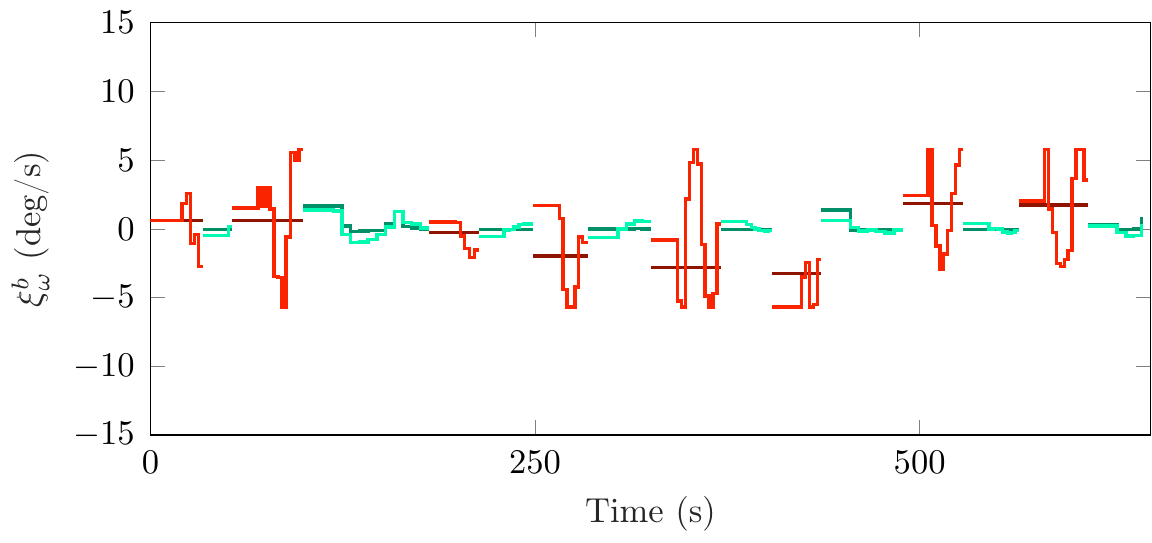}};
    	\node[draw=none,rectangle,minimum width=\columnwidth,minimum height=0.75in,anchor=south west] (A) at ($(B.north west)+(+0mm,0in)$)
    {\includegraphics[width=\columnwidth]{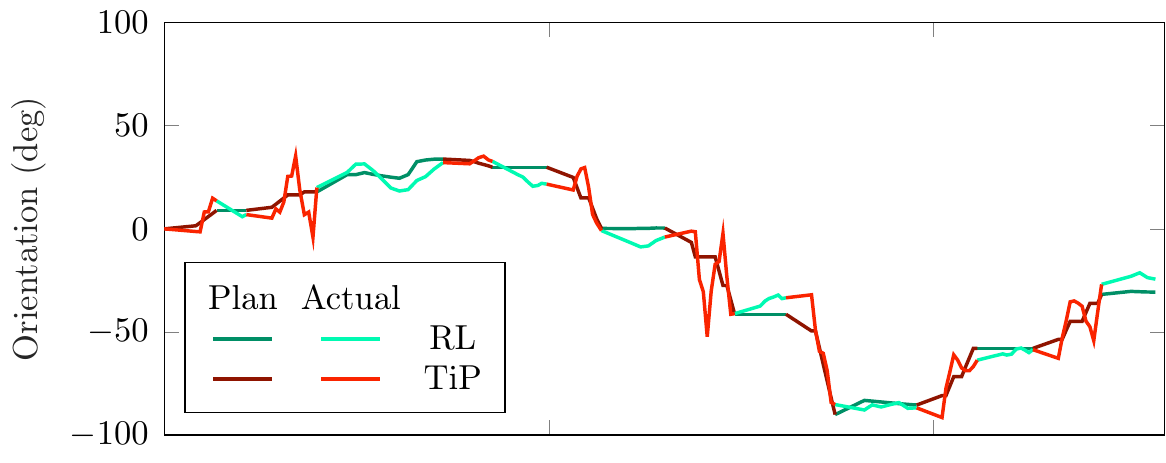}};    
    \node[draw=none,rectangle,minimum width=\columnwidth,minimum
    height=0.75in,anchor=south west,xshift=5pt] (D) at (B.south east)
    {\includegraphics[width=0.98\columnwidth]{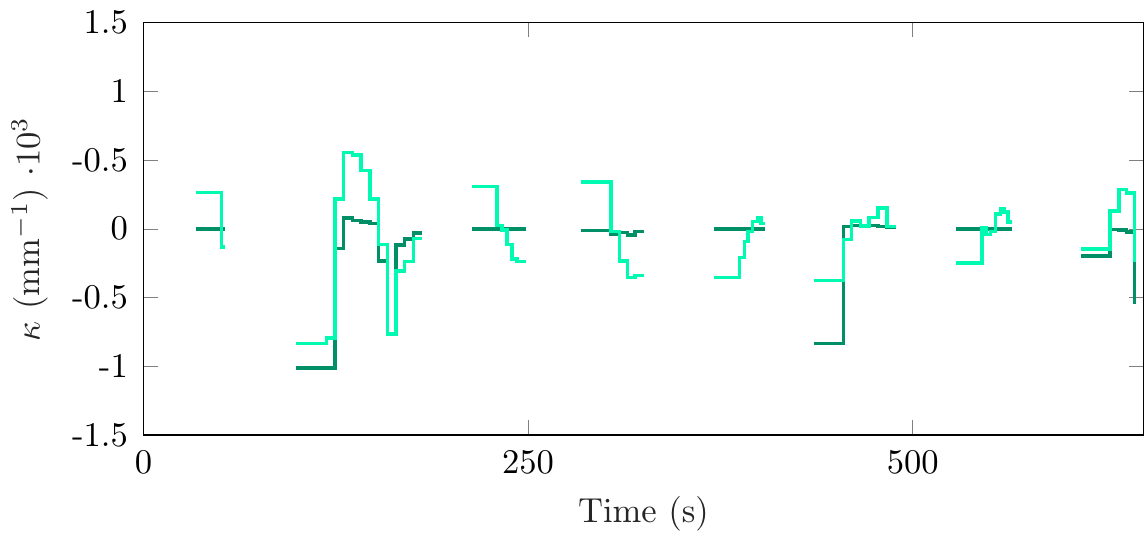}};
    \node[draw=none,rectangle,minimum width=\columnwidth,minimum
    height=0.75in,anchor=south west,xshift=5pt] (C) at ($(A.south east)+(-1.0mm,0in)$)
    {\includegraphics[width=\columnwidth]{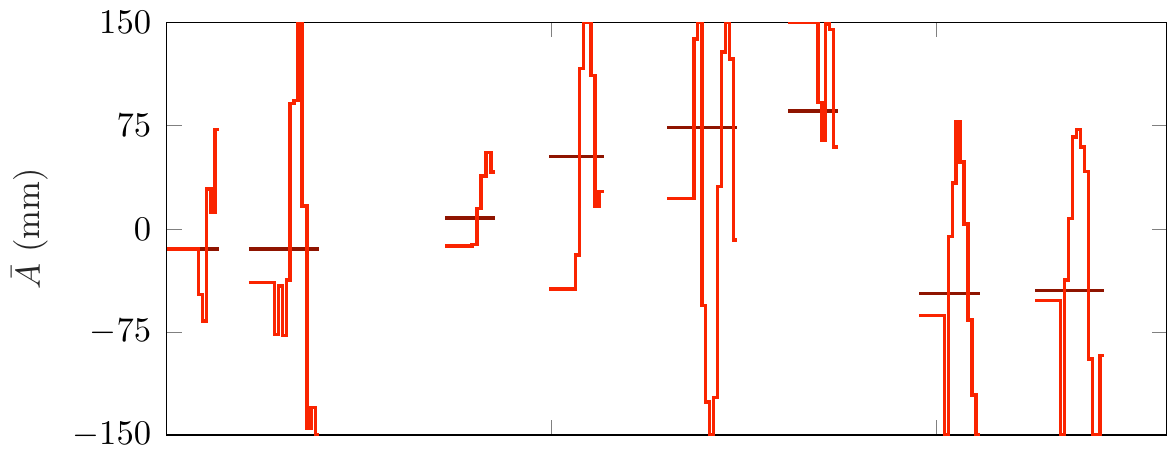}};
  \end{tikzpicture}}
  \vspace{-2.0mm}
  \subcaption{Scenario 2 signals}
  \vspace{3.0mm}
  \centering
  \scalebox{0.95}{
  \begin{tikzpicture}[inner sep=0pt, outer sep=0pt]
    \node[draw=none,rectangle,minimum width=\columnwidth,minimum height=0.75in,anchor=south west] (B) at (0in,0in)
    {\includegraphics[width=\columnwidth]{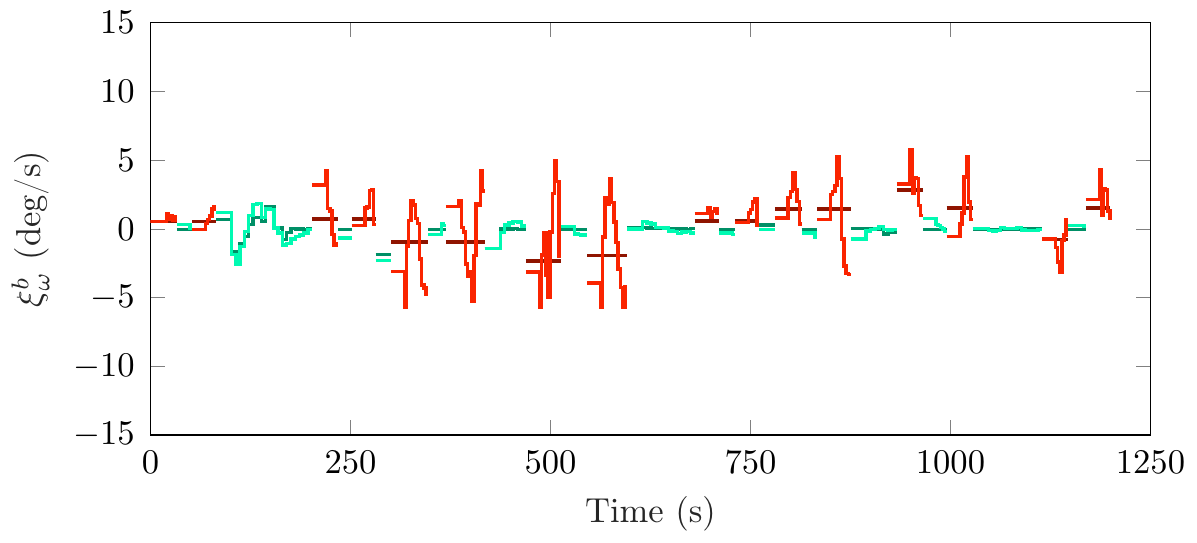}};
    	\node[draw=none,rectangle,minimum width=\columnwidth,minimum height=0.75in,anchor=south west] (A) at ($(B.north west)+(-1.5mm,0in)$)
    {\includegraphics[width=0.98\columnwidth]{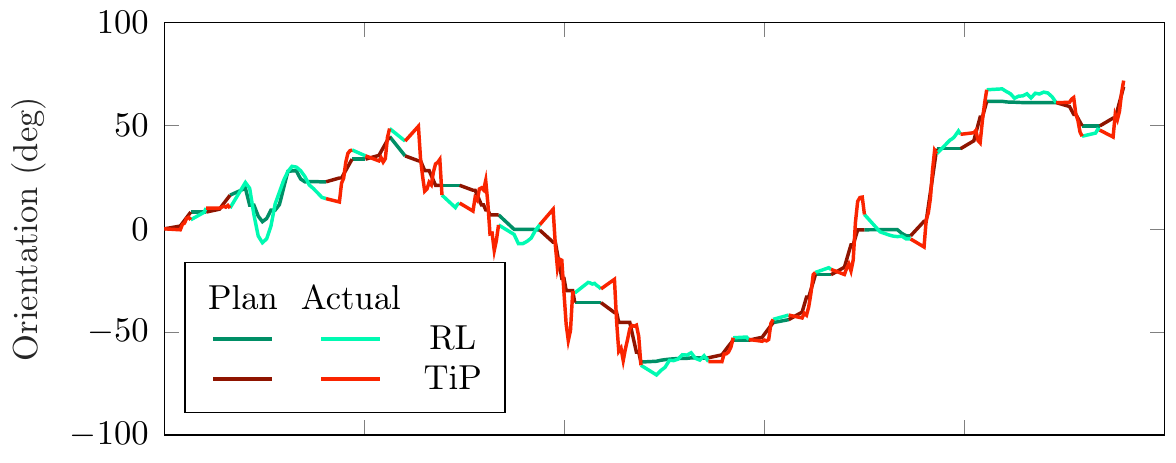}};
    \node[draw=none,rectangle,minimum width=\columnwidth,minimum
    height=0.75in,anchor=south west,xshift=5pt] (D) at (B.south east)
    {\includegraphics[width=\columnwidth]{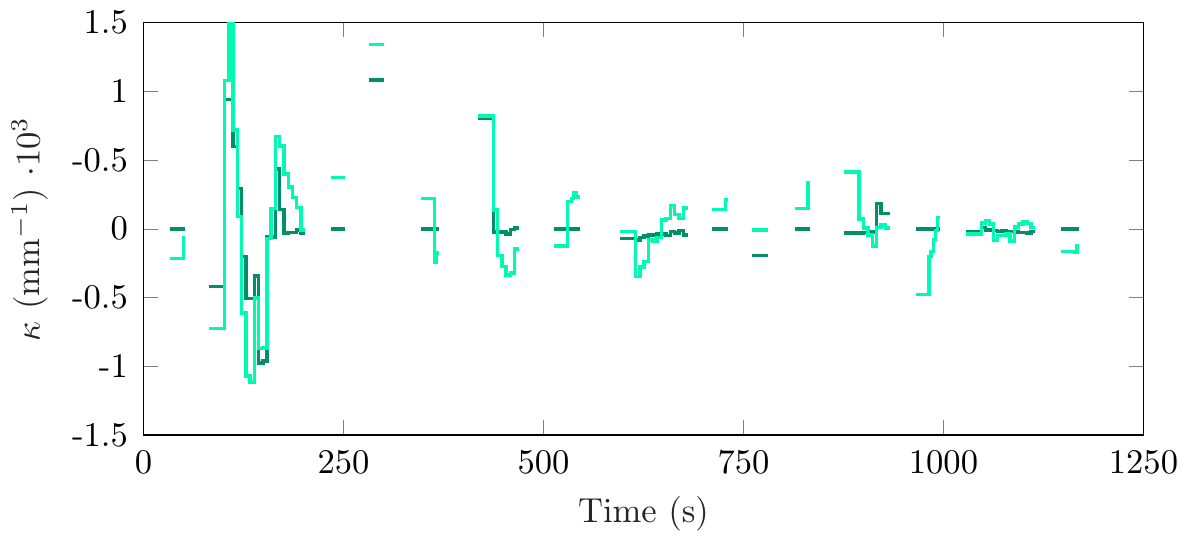}};
    \node[draw=none,rectangle,minimum width=\columnwidth,minimum
    height=0.75in,anchor=south west,xshift=5pt] (C) at ($(A.south east)+(-0.5mm,0in)$)
    {\includegraphics[width=0.99\columnwidth]{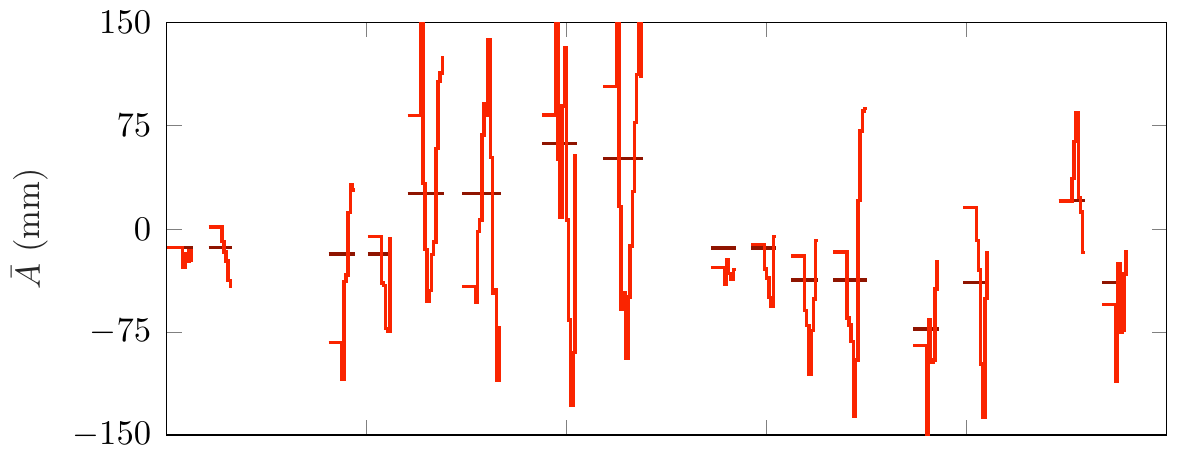}};
  \end{tikzpicture}}
  \vspace{-2.0mm}
  \subcaption{Scenario 3 signals.}
  \end{center}
  \caption{Time-varying signals describing Scenarios 1-3 in Figure
  \ref{fig:mult-gait-scen}: \textbf{(top-left)} orientation, $\theta(t)$
  \textbf{(bottom-left)} angular velocity, $\xi^b_{\omega}(t)$
  \textbf{(top-right)} Rectilinear curvature, $\kappa(t)$
  \textbf{(bottom-right)} Turn-in-Place peak amplitude, $\bar{A}(t)$.
  \label{signals}}
  \vspace{-10.0mm}
\end{figure*}

The control and orientation signals are plotted in Figure \ref{signals}.
Their piecewise nature is a function of the active gait changing as a
function of time, and therefore the applied control for the gait.
Whereas Figure \ref{fig:mult-gait-scen} confirms trajectory tracking,
Figure \ref{signals} confirmed orientation tracking (top left graph for
each scenario). Closed-loop discrete control under the empirically
regressed models successfully tracks the synthesized trajectories to
arrive at the goal point.


%% file: conclusion.tex

The paper described Turn-in-Place gait for snake-like robots, whose
standing wave and body-contact properties were engineered to primarily
produce net rotational motion. The rigid dynamics of the body frame were
derived using a shape-centric modeling framework
\cite{ChVe_RAS[2020]} with viscous body-ground friction.
Validation of the resulting integro-differential dynamical equations
employed the Gazebo simulator and physical experiments for a $12$-link
articulated robot.

The Turn-in-Place gait is the fourth motion primitive whose evolution
the shape-centric modeling approach is capable of describing. It is
further evidence that the framework constitutes a unified modeling
approach for describing the locomotive behavior of a variety of
gaits for snake-like robots.  In characterizing each gait's
time-averaged movement, we obtain a reduced-order control-to-action map
describing each gait's steady-state locomotive behavior as a function of
its input shape parameters. The maps resemble unicycle-like motion
profiles such that planning and control frameworks commonly used for
differential-drive vehicles apply to the snake-like robot.  Pairing the
Turn-in-Place gait with a previously modeled Rectilinear gait in a
multi-gait planning and tracking strategy permits traversal through
environments requiring high angular displacements.  The snake-like robot
successfully tracks multi-gait trajectory plans, in Gazebo simulation,
to complete several navigation scenerios.

The Turn-in-Place gait fills a critical gap in the space of motion
profiles achievable by snake-like robots. The utility of rotating
in-place is evident in the locomotion scenarios demonstrated here. By
filling this mobility gap, a greater breadth of planning and tracking
control approaches, commonly used for differential-drive vehicles,
become applicable for snake-like robot locomotion.  Future work aims to
replicate the simulations on the physical robot, once a means to track
its pose over long distances has been established.